\documentclass[letterpaper]{article} 
\usepackage{aaai2027}  
\usepackage[hyphens]{url}  
\usepackage{graphicx} 
\usepackage{natbib}  
\usepackage{caption} 
\usepackage{algorithm}
\usepackage{algorithmic}

\usepackage{newfloat}
\usepackage{listings}
\DeclareCaptionStyle{ruled}{labelfont=normalfont,labelsep=colon,strut=off} 
\floatstyle{ruled}
\newfloat{listing}{tb}{lst}{}
\floatname{listing}{Listing}

\usepackage{booktabs}

\usepackage{booktabs}
\usepackage{multirow}
\usepackage{graphicx}
\usepackage[table,dvipsnames]{xcolor}
\usepackage{makecell}

\definecolor{GainGreen}{HTML}{169B45}
\definecolor{LossRed}{HTML}{D64545}
\definecolor{PanelBlue}{HTML}{EAF2FF}
\definecolor{HeaderBlue}{HTML}{F5F8FD}
\definecolor{OursBlue}{HTML}{F8FAFF}
\definecolor{NoteGray}{HTML}{667085}
\definecolor{GainGreen}{HTML}{169B45}
\definecolor{LossRed}{HTML}{D64545}
\definecolor{PanelBlue}{HTML}{EAF2FF}
\definecolor{HeaderBlue}{HTML}{F5F8FD}
\definecolor{OursBlue}{HTML}{F8FAFF}
\definecolor{NoteGray}{HTML}{667085}

\newcommand{\scoregain}[1]{%
  \hspace{0.12em}%
  {\scriptsize\bfseries\textcolor{GainGreen}{$\uparrow$\,#1}}%
}

\newcommand{\scoreloss}[1]{%
  \hspace{0.12em}%
  {\scriptsize\bfseries\textcolor{LossRed}{$\downarrow$\,#1}}%
}

\usepackage{pifont}

\newcommand{\cmark}{\ding{51}}
\newcommand{\xmark}{\ding{55}}

\newcommand{\RcbIcon}[1]{%
  \IfFileExists{Figures/logos/#1.png}{%
    \raisebox{-0.13em}{%
      \includegraphics[
        draft=false,
        height=0.90em,
        keepaspectratio
      ]{Figures/logos/#1.png}%
    }%
  }{%
    \PackageWarning{MainTable}{Icon Figures/logos/#1.png not found}%
    \makebox[0.90em][c]{\scriptsize ?}%
  }%
  \hspace{0.20em}%
}

\newcommand{\OpenAIIcon}{\RcbIcon{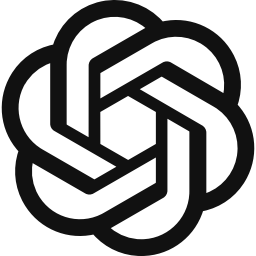}}
\newcommand{\GlmIcon}{\RcbIcon{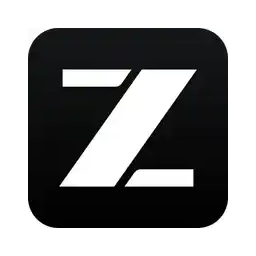}}
\newcommand{\MiniMaxIcon}{\RcbIcon{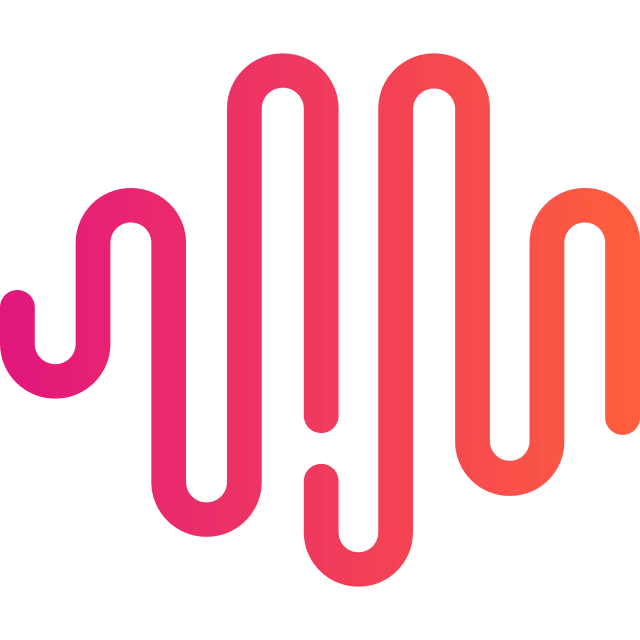}}
\newcommand{\DeepSeekIcon}{\RcbIcon{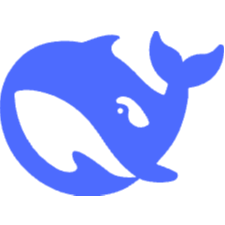}}

\newcommand{\CodexIcon}{\RcbIcon{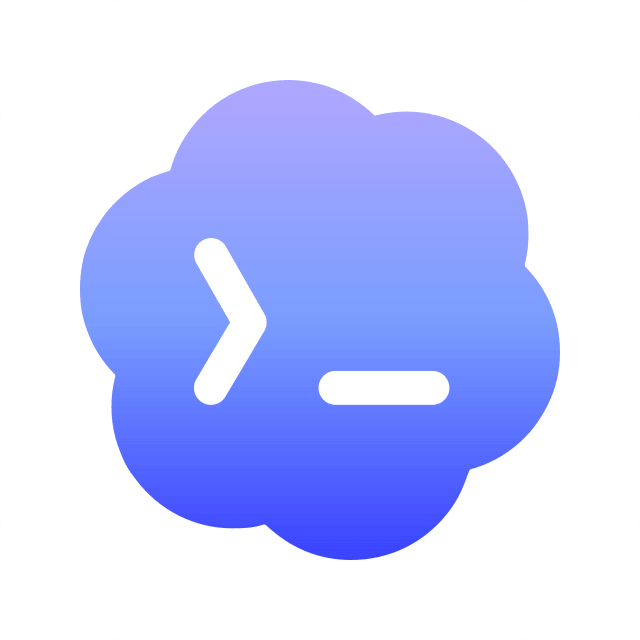}}
\newcommand{\ClaudeIcon}{\RcbIcon{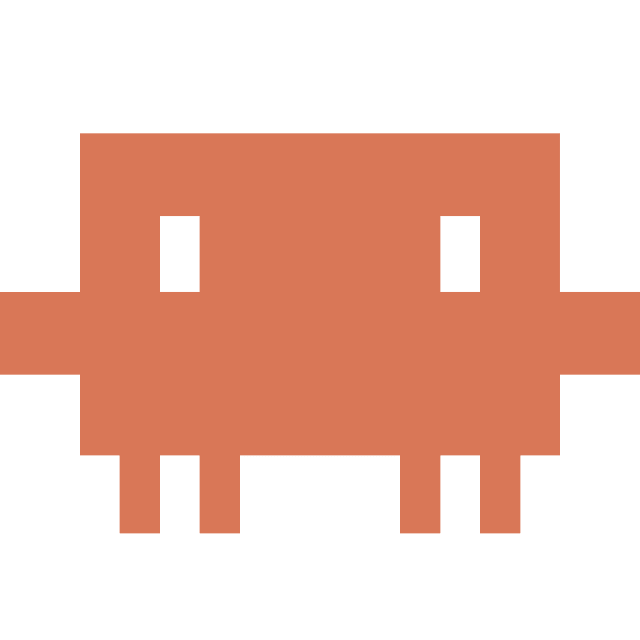}}
\newcommand{\OpenClawIcon}{\RcbIcon{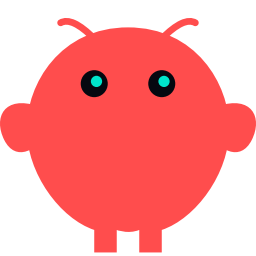}}
\newcommand{\OpenScienceIcon}{\RcbIcon{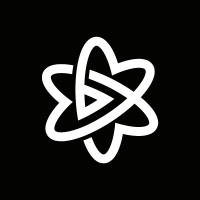}}

\newcommand{\ModelCell}[2]{%
  \cellcolor{white}%
  \multirow{-2}{*}[-0.10ex]{%
    \begingroup
    \setlength{\fboxsep}{0pt}%
    \colorbox{white}{\strut #1#2}%
    \endgroup
  }%
}

\definecolor{OverallHeaderBlue}{HTML}{DCE8F5}
\definecolor{OverallVanillaBlue}{HTML}{F2F6FA}
\definecolor{OverallOursBlue}{HTML}{E2EDF8}

\newcommand{\OverallHeader}{%
  \cellcolor{OverallHeaderBlue}\textbf{Overall}%
}

\newcommand{\OverallVanilla}[1]{%
  \cellcolor{OverallVanillaBlue}#1%
}

\newcommand{\OverallOurs}[2]{%
  \cellcolor{OverallOursBlue}%
  \textbf{#1}\scoregain{#2}%
}

\newcommand{\our}{AutoSciRub}

\usepackage[most]{tcolorbox}
\usepackage{listings}
\usepackage{xcolor}

\definecolor{PromptFrame}{HTML}{6F927D}
\definecolor{PromptBackground}{HTML}{EEF7F1}

\newtcblisting{PromptBox}[1]{
  enhanced,
  breakable,
  listing only,
  title={#1},
  colback=PromptBackground,
  colframe=PromptFrame,
  colbacktitle=PromptFrame,
  coltitle=white,
  fonttitle=\bfseries\small,
  boxrule=0.7pt,
  arc=2pt,
  left=2mm,
  right=2mm,
  top=1.5mm,
  bottom=1.5mm,
  before skip=8pt,
  after skip=10pt,
  listing options={
    basicstyle=\rmfamily\footnotesize,
    numbers=none,
    numberstyle=\scriptsize,
    breaklines=true,
    breakatwhitespace=true,
    breakautoindent=false,
    breakindent=0pt,
    columns=fullflexible,
    keepspaces=true,
    showspaces=false,
    showstringspaces=false,
    showtabs=false,
    tabsize=2,
    xleftmargin=0pt,
    frame=none,
    aboveskip=0pt,
    belowskip=0pt
  }
}
\usepackage{newfloat}
\usepackage{listings}
\DeclareCaptionStyle{ruled}{labelfont=normalfont,labelsep=colon,strut=off} 
\floatstyle{ruled}
\newfloat{listing}{tb}{lst}{}
\floatname{listing}{Listing}

\usepackage{booktabs}
\usepackage{graphicx}
\usepackage{subcaption}

\title{Learning to Evaluate Before Improving: Automatic Rubric Induction for Automatic Research Agents}
\author {
    Xuehai Wang\textsuperscript{\rm 1},
    Haowei Qin\textsuperscript{\rm 2},
    Tongxin Liu\textsuperscript{\rm 3},
    Junkai Li\textsuperscript{\rm 4},
    Buqiang Xu\textsuperscript{\rm 1},
    Jintian Zhang\textsuperscript{\rm 1},
    Yijun Chen\textsuperscript{\rm 1},
    Zirui Xue\textsuperscript{\rm 1},
    Shumin Deng\textsuperscript{\rm 1}\corresponding
}
\affiliations {
    \textsuperscript{\rm 1}Zhejiang University \\
    \textsuperscript{\rm 2}University of Electronic Science and Technology of China\\
    \textsuperscript{\rm 3}Beijing University of Posts and Telecommunications\\
    \textsuperscript{\rm 4}Zhejiang University of Technology\\
    22651308@zju.edu.cn, 231sm@zju.edu.cn
}

\begin{document}
\nocopyright
\maketitle






\begin{abstract}
Autonomous scientific research agents are increasingly applied to end-to-end scientific workflows, including literature review, data analysis, experimentation, and report generation. However, open-ended research tasks often do not clearly specify the analyses, methods, and success criteria required to complete the task. As a result, agents may miss important analyses, use inappropriate methods, or draw conclusions that are insufficiently supported by evidence.
To address the problem, we present \textbf{AutoSciRub}, an evaluation-first framework that induces a task-specific executable rubric before research execution, and uses it to guide execution, criterion-level verification as well as iterative revision. AutoSciRub decomposes an underspecified instruction into atomic scientific goals, grounds them in relevant literature and task-visible data, and synthesizes specific, actionable, and verifiable criteria. The resulting rubric makes implicit experimental and evidential requirements explicit, providing  guidance for experiments and analyses. During revision, rubric-guided verification identifies unmet criteria and enables targeted refinement of the research report and its supporting artifacts. 
On ResearchClawBench, AutoSciRub consistently improves all tested configurations, with an average gain of \textbf{2.08} points across three backbone LLMs under the fixed Codex harness and \textbf{2.95} points across three agent harnesses using a fixed DeepSeek-V4-Flash backbone. On a randomly sampled 20-task subset of AstaBench E2E Discovery, AutoSciRub further achieves an average improvement of \textbf{16.8} points across three agent harnesses, while maintaining or increasing the number of successfully completed tasks. These results demonstrate that evaluation-first guidance provides an effective and generalizable control mechanism for autonomous scientific research\footnote{The code is available at \url{https://github.com/zjunlp/AutoSciRub}.}.
\end{abstract}

\section{Introduction}
\label{Introduction}

\begin{figure}[!t]
  \centering
  \includegraphics[width=\linewidth]{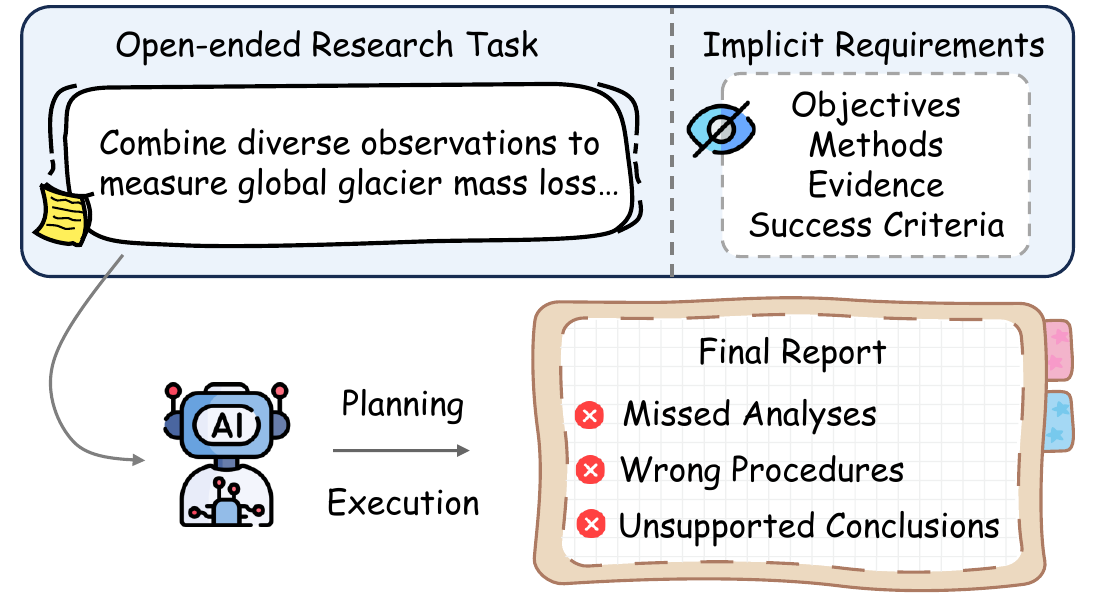}
  \caption{
      Why autonomous research agents struggle with open-ended research tasks.
  }
  \label{fig:intro}
\end{figure}

Large language model agents are increasingly being used to automate scientific research workflows, including literature review, hypothesis formulation, experimental design, code implementation, result analysis, and scientific report generation~\cite{agentic_science_survey}.
Recent systems have further advanced toward multi-agent and long-horizon research processes that produce both scientific reports and supporting experimental artifacts~\cite{ai_scientist,agentlaboratory,Kosmos}. 
However, as their execution capabilities improve, a fundamental challenge becomes more pressing: \emph{how can we ensure that the resulting research actually fulfills the scientific intent of the original instruction?} 
Scientific instructions often specify only a high-level objective while leaving intermediate goals, methodological requirements, expected evidence, and success conditions implicit. As a result, an agent may produce a plausible-looking report while omitting essential analyses, using inappropriate procedures, or making claims unsupported by experimental evidence. Figure~\ref{fig:intro} illustrates how these implicit requirements can lead to failures throughout research planning and execution.

Existing scientific-agent benchmarks evaluate research outputs using executable metrics, expert-authored rubrics, and LLM-based judgments~\cite{agent_evaluation_survey}. Expert-authored rubrics are particularly useful because they provide explicit and interpretable criteria for assessing complex research artifacts~\cite{PaperBench}. However, constructing them requires substantial domain expertise and manual effort, making them difficult to scale to new research tasks.
Recent work on automatic rubric generation offers a promising alternative by deriving task-specific criteria from natural-language instructions and contextual information~\cite{rubric_survey,GER_Eval,AdaRubric}. Nevertheless, scientific rubric generation is especially challenging because valid criteria depend not only on the instruction, but also on task-relevant literature, available data, domain conventions, and execution constraints. Without such grounding, automatically generated criteria may be incomplete, scientifically unjustified, or infeasible to verify.

Moreover, rubrics in existing scientific-agent benchmarks are predominantly treated as \emph{post-hoc} evaluation instruments. They score completed artifacts, but do not help the agent determine what evidence should be produced or how an incomplete artifact should be improved. We argue that rubrics should instead serve as \emph{intermediate scientific specifications}. They should make the implicit requirements of a research task explicit and connect task interpretation, research execution, output verification, and iterative revision. Under this view, a reliable research agent should \textit{\textbf{learn to evaluate before improving}}.

Based on this principle, we introduce \textbf{AutoSciRub}, a framework that automatically induces evidence-grounded scientific rubrics and uses them to guide iterative research improvement. 
AutoSciRub consists of two stages. 
First, \emph{Automatic Rubric Induction} decomposes the original instruction into atomic scientific goals and grounds them in relevant literature, web evidence, available task data, and environmental constraints. The resulting task-specific rubric makes the required analyses, evidence, and success conditions explicit. 
Second, after the agent produces a research report and supporting artifacts, \emph{Rubric-Guided Iterative Revision} evaluates the available evidence criterion by criterion, identifies unmet scientific requirements, and provides targeted feedback for revision.

We evaluate AutoSciRub on all 40 tasks in ResearchClawBench and a fixed subset of 20 end-to-end scientific discovery tasks from AstaBench.
On ResearchClawBench, AutoSciRub yields an average improvement of \textbf{2.08} points across three backbone LLMs under the fixed Codex harness and \textbf{2.95} points across three agent harnesses using the fixed DeepSeek-V4-Flash backbone.
On AstaBench, it further improves scores by an average of \textbf{16.8} points across three agent systems.
These results demonstrate that automatically induced, evidence-grounded rubrics can provide scalable evaluation criteria while also serving as effective guidance for autonomous scientific research.


\section{Related Work}

\textbf{Autonomous Scientific Research Agents and Benchmarks.}
Early systems organize end-to-end scientific research as single-agent pipelines or centrally orchestrated workflows, coordinating literature review, experimentation, and manuscript preparation within a unified process~\cite{ai_scientist,ai_scientist_v2,data2paper}.
Multi-agent systems instead distribute these stages across specialized roles for collaborative ideation, critical discussion, experiment execution, and research coordination~
\cite{agentlaboratory,Robin,AI-Researcher,accelerating}, while shared research states support long-horizon exploration, failure recovery, and iterative hypothesis refinement~
\cite{Kosmos,EvoScientist,AutoScientists,AutoResearchClaw,Structured_Episodic_Event_Memory}.
Complementary memory and context-management systems further support persistent agents through structured long-horizon memory~\cite{structmem,lycheememoryv2}, cache-efficient context control~\cite{tokenpilot}, and hierarchical multimodal experience storage~\cite{lightmemego}.
In parallel, scientific-agent benchmarks have progressed from executable coding and model-development tasks under fixed objectives~
\cite{ScienceAgentBench,MLE-bench}, to research reproduction that requires reconstructing methods and experiments~\cite{PaperBench,xKG}, and further to end-to-end settings that assess experiment formulation, raw-evidence analysis, and scientific report generation from high-level instructions~
\cite{AstaBench,ResearchGym,FIREBench,ResearchClawBench}.

\textbf{Rubric-Guided Evaluation and Iterative Refinement.}
Rubric-based evaluation offers a structured alternative to holistic judging through explicit, fine-grained criteria~\cite{LLM_Rubric,rubric_survey}.
Expert-authored hierarchical rubrics decompose complex research artifacts into independently assessable components~\cite{PaperBench,ResearchRubrics}, while later benchmarks introduce atomic, verifiable criteria and question-specific scoring or deduction points~\cite{DeepResearch_Bench2,SedarEval}.
Rubric construction has become increasingly automated~\cite{GER_Eval,AdaRubric,ThinkWithRubrics}.
Recent methods generate dataset- or instance-specific criteria~\cite{dynamic_rubric}, retrieve rubric knowledge from related queries~\cite{RubricRAG,chen2026automated}, or recursively expand questions into structured criteria trees~\cite{Qworld,DeepRubric}; recursive decomposition and filtering further reduce redundancy, overlap, and preference misalignment~\cite{Rethinking_Rubric,hong2026can}.
Recent refinement frameworks combine explicit rubrics with test-time verifiers or pre-execution checks to assess outputs against task-specific criteria~\cite{wan2026inference,RubricRefine,Evaluation_driven,can_we_pre}.
Criterion-level diagnoses are then fed back into generation for targeted revision of responses~\cite{Self-Refine,CRITIC}, agent trajectories, and tool-use programs~\cite{RubricRefine}.
\section{Method}

\begin{figure*}[!t]
  \centering
  \includegraphics[width=\linewidth]{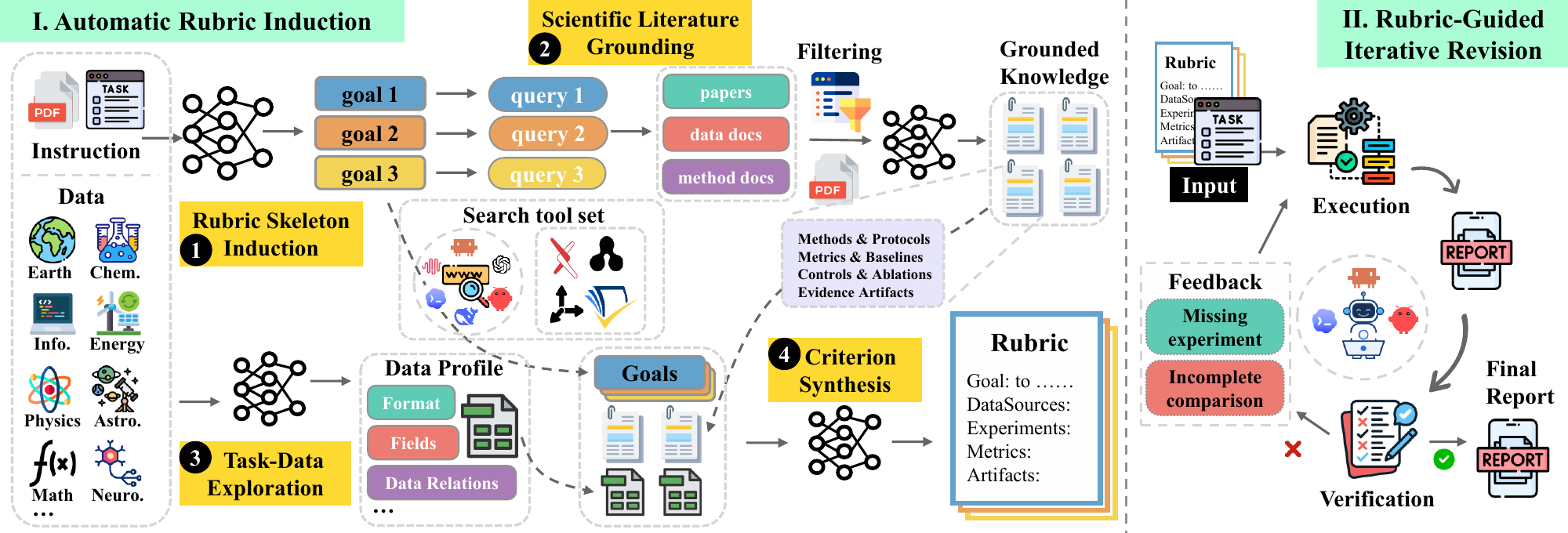}
  \caption{
Overview of AutoSciRub. (1) Automatic Rubric Induction: starting from a multi-domain scientific task, AutoSciRub induces a rubric skeleton, grounds it in scientific literature, explores the task-visible data, and synthesizes a task-specific executable rubric. (2) Rubric-Guided Iterative Revision: the induced rubric guides research execution and supports an iterative feedback loop that verifies and improves the generated artifact into the final report.
}
  \label{fig:method}
\end{figure*}

Autonomous research agents often revise their outputs without an explicit specification of the scientific requirements that the task entails. Our framework addresses this limitation by first transforming an underspecified research instruction into a task-specific executable rubric and then using the rubric to guide research execution, verification, and revision. As illustrated in Figure~\ref{fig:method}, Automatic Rubric Induction first constructs an instruction-derived rubric skeleton through Rubric Skeleton Induction, and then transforms it into a task-specific executable rubric through Scientific Literature Grounding, Task-Data Exploration, and Criterion Synthesis.
Rubric-Guided Iterative Revision subsequently evaluates the generated research artifact criterion by criterion and provides targeted feedback on missing experiments, comparisons, and evidence.  We first formulate the research task and then describe the two stages in detail.

\subsection{Task Formulation}
\label{sec:task_formulation}

We consider a collection of multi-domain, end-to-end autonomous scientific research tasks $\mathcal{D}$ spanning
\emph{Astronomy}, \emph{Chemistry}, \emph{Earth Science},
\emph{Energy Science}, \emph{Information Science}, \emph{Life Science},
\emph{Materials Science}, \emph{Mathematics}, \emph{Neuroscience}, and
\emph{Physics}. Each task is represented as
$\tau_i=(x_i,\mathcal{E}_i)$, where $x_i$ is a high-level research instruction and $\mathcal{E}_i$ is the task-visible environment, including available literature, data, search, code execution, and domain-specific tools.

Given $\tau_i$, a research agent $\pi$ performs scientific reasoning, experimentation, and analysis to produce
\begin{equation}
\mathcal{A}_i^\pi
=
\pi(x_i,\mathcal{E}_i)
=
(r_i,\mathcal{S}_i),
\label{eq:research_artifact}
\end{equation}
where $r_i$ is the scientific report and $\mathcal{S}_i$ contains its supporting code, results, analyses, tables, and figures.

Research instructions are typically underspecified: they state a broad objective without fully defining the scientific questions, comparisons, or evidence required. We therefore associate each task with a latent scientific specification
\begin{equation}
\mathcal{Z}_i^*
=
(\mathcal{G}_i^*,\mathcal{C}_i^*),
\label{eq:latent_specification}
\end{equation}
where $\mathcal{G}_i^*$ denotes the required scientific goals and $\mathcal{C}_i^*$ the evidence requirements for verifying them. Artifact quality is defined by
\begin{equation}
q_i^\pi
=
\operatorname{Eval}
\left(
\mathcal{A}_i^\pi;
\mathcal{Z}_i^*
\right),
\label{eq:artifact_quality}
\end{equation}
where $\operatorname{Eval}(\cdot)$ assesses goal coverage and evidence adequacy. Our objective is to improve the artifact quality across tasks.

Since neither $\mathcal{Z}_i^*$ nor its evaluator is available during execution, AutoSciRub induces an \emph{instruction-derived rubric skeleton} $\widehat{\mathcal{G}}_i$ and a \emph{task-specific executable rubric} $\widehat{\mathcal{R}}_i$ from the instruction, scientific literature, and task-visible data. The rubric operationalizes the latent specification and guides execution, criterion-level verification, and
iterative revision.

\subsection{Automatic Rubric Induction}
\label{sec:rubric_induction}

Rather than selecting criteria from a fixed library, we induce a task-specific executable rubric at inference time. As shown in Figure~\ref{fig:method}, this process consists of four steps: Rubric Skeleton Induction, Scientific Literature Grounding, Task-Data Exploration, and Criterion Synthesis. These steps make the instruction's implicit requirements explicit, ground them in established scientific practice and task-visible resources, and convert them into verifiable
criteria.

\paragraph{Rubric Skeleton Induction.}

A high-level research instruction often implies multiple analyses, comparisons, and interpretations without listing them explicitly. We organize these requirements into a compact set of atomic scientific goals:
\begin{equation}
\widehat{\mathcal{G}}_i
=
\phi_{\mathrm{inst}}(x_i)
=
\left\{
g_{i,k}
\right\}_{k=1}^{K_i},
\label{eq:rubric_skeleton}
\end{equation}
where $K_i$ is the number of induced goals and $g_{i,k}=(n_{i,k},s_{i,k})$ contains a concise goal name $n_{i,k}$ and a concrete scientific requirement $s_{i,k}$. Together, these goals form an instruction-derived rubric skeleton.

The goals must be traceable to the instruction, cover its main requirements, and avoid unnecessary overlap. At this stage, the agent observes only $x_i$ and does not introduce specific methods, metrics, baselines, or expected results. The skeleton therefore defines
\emph{what} the task should address, while later steps determine \emph{how} each goal should be established.

\paragraph{Scientific Literature Grounding.}

The rubric skeleton defines the task scope but not the scientific practice needed to satisfy each goal. For every $g_{i,k}$, the agent forms broad queries over relevant concepts, methods, metrics, and standard protocols. It first consults task-provided related literature and then supplements missing coverage through the native web-search capability of the agent harness and external services, including arXiv, OpenAlex, Semantic Scholar, and Tavily. The retrieved sources may include papers, preprints, official dataset or benchmark pages, method repositories, and software documentation.

Queries avoid long instruction spans, task identifiers, and task-specific filenames. Candidate papers are ranked by relevance to the induced goals. Before full-text access, the retrieval layer removes candidates whose titles match a hidden target-paper blocklist maintained by the evaluation harness and unavailable to the induction agent. The agent then retains five to seven core papers for the task and consults additional documentation when needed.

For each goal, the selected sources are summarized as
\begin{equation}
\mathcal{K}_{i,k}
=
\phi_{\mathrm{lit}}
\left(
g_{i,k};
\widetilde{\mathcal{L}}_i
\right),
\label{eq:literature_grounding}
\end{equation}
where $\widetilde{\mathcal{L}}_i$ is the filtered source set. $\mathcal{K}_{i,k}$ records relevant methods and protocols, typical analyses, metrics, baselines, controls, ablations, robustness checks, and useful evidence forms. The information is organized by scientific goal rather than by individual paper. Retrieved sources guide experiment and evidence design but are not treated as experimental evidence for the
current task.

\paragraph{Task-Data Exploration.}

Literature grounding describes what is scientifically appropriate, but the proposed analyses must also be supported by the available resources. The agent therefore performs a lightweight inspection of the task-visible data:
\begin{equation}
\mathcal{P}_i
=
\phi_{\mathrm{data}}
\left(
\mathcal{E}_i
\right),
\label{eq:task_data_profile}
\end{equation}
where $\mathcal{P}_i$ records the available files and datasets, their formats and dimensions, key fields, units, labels, experimental conditions, cross-source relations, and relevant constraints.

This profile identifies which data sources support each goal and which literature-suggested analyses are feasible without assuming unavailable labels, conditions, or reference values. It is used only for planning; full experiments and analyses are deferred to task execution.

\paragraph{Criterion Synthesis.}

Finally, the agent combines the rubric skeleton, literature-grounded knowledge, and task-data profile. It selects feasible experiments and analyses for each goal and specifies the evidence needed to support the resulting claims. These requirements are synthesized into a task-specific executable rubric:
\begin{equation}
\widehat{\mathcal{R}}_i
=
\phi_{\mathrm{syn}}
\left(
\widehat{\mathcal{G}}_i,
\left\{
\mathcal{K}_{i,k}
\right\}_{k=1}^{K_i},
\mathcal{P}_i
\right)
=
\left\{
\rho_{i,j}
\right\}_{j=1}^{M_i},
\label{eq:grounded_rubric}
\end{equation}
where $M_i$ is the number of synthesized criteria. Each criterion $\rho_{i,j}$ links to one or more scientific goals and specifies the relevant data sources, required experiments or analyses, evaluation metrics and comparisons, expected evidence artifacts, and its
satisfaction condition.

The resulting rubric aligns established scientific practice with the evidence that can be produced from the current task, providing an explicit specification for subsequent research execution and verification.

\subsection{Rubric-Guided Iterative Revision}
\label{sec:iterative_revision}

The induced rubric is provided to the backbone research agent as an execution-time specification. Conditioned on $\widehat{\mathcal{R}}_i$, the agent performs the required experiments and analyses and produces an initial artifact $\mathcal{A}_i^{(0)}=(r_i^{(0)},\mathcal{S}_i^{(0)})$, containing the report and its supporting code, results, tables, and figures.

\paragraph{Criterion-Level Verification.}

The initial execution may still leave some rubric requirements unmet.
At revision round $t$, a verifier checks each criterion against the current artifact:
\begin{equation}
\left(
z_{i,j}^{(t)},d_{i,j}^{(t)}
\right)
=
\operatorname{Verify}
\left(
\rho_{i,j},\mathcal{A}_i^{(t)}
\right),
\label{eq:criterion_verification}
\end{equation}
where $z_{i,j}^{(t)}\in\{0,1\}$ indicates whether the criterion is satisfied, and $d_{i,j}^{(t)}$ describes the remaining evidence gap.
The verifier considers the required experiments, reported results, supporting artifacts, and whether the conclusions are adequately supported. This criterion-level check identifies specific omissions rather than returning only a holistic score.

\paragraph{Targeted Revision.}

Failed criteria and their diagnoses form the revision feedback $\Delta_i^{(t)}$. The agent then updates the current artifact by addressing these gaps:
\begin{equation}
\mathcal{A}_i^{(t+1)}
=
\pi_{\mathrm{rev}}
\left(
\mathcal{A}_i^{(t)},
\Delta_i^{(t)},
\mathcal{E}_i
\mid
\widehat{\mathcal{R}}_i
\right).
\label{eq:targeted_revision}
\end{equation}
A revision may add missing experiments or comparisons, correct evidence artifacts, strengthen the analysis, or remove unsupported claims. The process stops when all criteria are satisfied or the revision budget is reached, yielding the final artifact $\mathcal{A}_i^{*}=\mathcal{A}_i^{(T)}$.

\begin{figure}[!t]
  \centering
  \includegraphics[width=\linewidth]{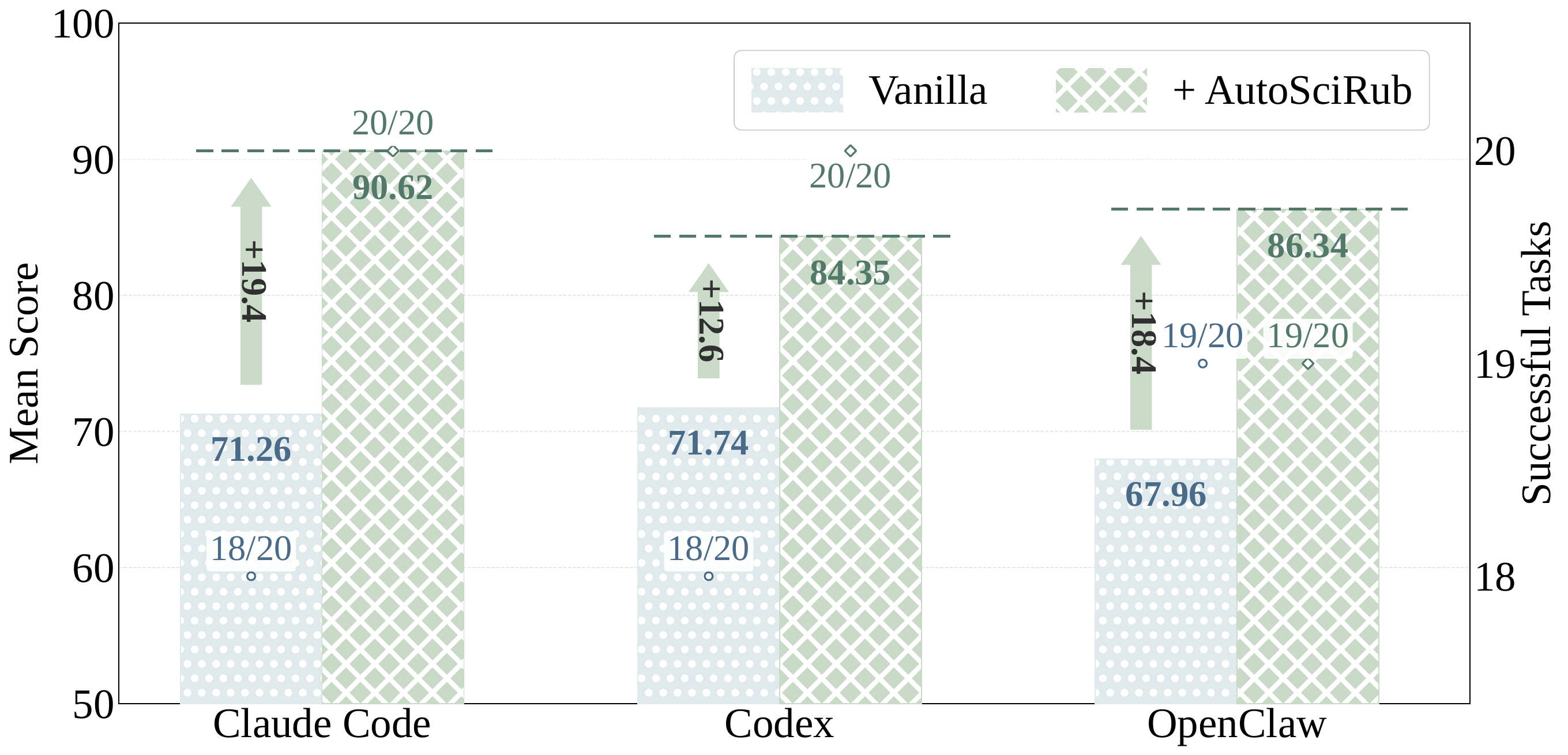}
  \caption{
 Performance comparison on AstaBench.
      AutoSciRub consistently improves mean benchmark scores across all three evaluated agent configurations.
  }
  \label{fig:astabench_results}
\end{figure}
\section{Experiments}
\begin{table*}[t]
\centering

\begingroup
\footnotesize

\setlength{\tabcolsep}{3.4pt}
\renewcommand{\arraystretch}{1.17}

\resizebox{\textwidth}{!}{%
\begin{tabular}{
@{}
ll
c
cccccccccc
@{}
}

\toprule

\rowcolor{PanelBlue}
\multicolumn{13}{@{}l@{}}{%
  \textbf{A. Cross-Model Generalization}
  \hspace{0.6em}
  {\normalfont\itshape\textcolor{NoteGray}{%
    Fixed agent harness: \CodexIcon Codex%
  }}%
}
\\

\rowcolor{HeaderBlue}
\textbf{Backbone LLM}
& \textbf{Setting}
& \OverallHeader
& \textbf{Astro.}
& \textbf{Chem.}
& \textbf{Earth}
& \textbf{Energy}
& \textbf{Info.}
& \textbf{Life}
& \textbf{Mater.}
& \textbf{Math}
& \textbf{Neuro.}
& \textbf{Phys.}
\\

\midrule

& \textit{vanilla}
& \OverallVanilla{18.66}
& 29.69
& 10.97
& 21.25
& 17.87
& \textbf{17.61}
& 16.02
& 17.55
& 16.65
& 5.60
& 33.43
\\

\rowcolor{OursBlue}
\ModelCell{\OpenAIIcon}{GPT-5.4}
& {+\our}
& \OverallOurs{21.04}{2.38}
& \textbf{31.43}\scoregain{1.74}
& \textbf{11.62}\scoregain{0.65}
& \textbf{22.22}\scoregain{0.97}
& \textbf{22.70}\scoregain{4.83}
& 13.92\scoreloss{3.69}
& \textbf{19.85}\scoregain{3.83}
& \textbf{22.70}\scoregain{5.15}
& \textbf{20.57}\scoregain{3.92}
& \textbf{7.72}\scoregain{2.12}
& \textbf{37.69}\scoregain{4.26}
\\

\cmidrule(lr){1-13}

& \textit{vanilla}
& \OverallVanilla{20.86}
& 31.56
& 11.96
& \textbf{23.99}
& 17.95
& 15.69
& 16.88
& 22.14
& \textbf{24.20}
& 5.84
& 38.40
\\

\rowcolor{OursBlue}
\ModelCell{\GlmIcon}{GLM-5.2}
& {+\our}
& \OverallOurs{22.73}{1.87}
& \textbf{31.99}\scoregain{0.43}
& \textbf{17.85}\scoregain{5.89}
& 23.25\scoreloss{0.74}
& \textbf{20.70}\scoregain{2.75}
& \textbf{18.00}\scoregain{2.31}
& \textbf{17.34}\scoregain{0.46}
& \textbf{23.48}\scoregain{1.34}
& 23.03\scoreloss{1.17}
& \textbf{12.58}\scoregain{6.74}
& \textbf{39.06}\scoregain{0.66}
\\

\cmidrule(lr){1-13}

& \textit{vanilla}
& \OverallVanilla{19.05}
& \textbf{28.01}
& 13.36
& 19.14
& 19.65
& 9.31
& 15.10
& \textbf{22.31}
& \textbf{22.96}
& 6.30
& 34.37
\\

\rowcolor{OursBlue}
\ModelCell{\MiniMaxIcon}{MiniMax-M3}
& {+\our}
& \OverallOurs{21.04}{1.99}
& 27.10\scoreloss{0.91}
& \textbf{13.94}\scoregain{0.58}
& \textbf{21.53}\scoregain{2.39}
& \textbf{26.46}\scoregain{6.81}
& \textbf{12.20}\scoregain{2.89}
& \textbf{17.58}\scoregain{2.48}
& 20.86\scoreloss{1.45}
& 22.38\scoreloss{0.58}
& \textbf{12.25}\scoregain{5.95}
& \textbf{36.10}\scoregain{1.73}
\\

\addlinespace[3pt]
\midrule

\rowcolor{PanelBlue}
\multicolumn{13}{@{}l@{}}{%
  \textbf{B. Cross-Harness Generalization}
  \hspace{0.6em}
  {\normalfont\itshape\textcolor{NoteGray}{%
    Fixed backbone LLM:
    \DeepSeekIcon DeepSeek-V4-Flash%
  }}%
}
\\

\rowcolor{HeaderBlue}
\textbf{Agent Harness}
& \textbf{Setting}
& \OverallHeader
& \textbf{Astro.}
& \textbf{Chem.}
& \textbf{Earth}
& \textbf{Energy}
& \textbf{Info.}
& \textbf{Life}
& \textbf{Mater.}
& \textbf{Math}
& \textbf{Neuro.}
& \textbf{Phys.}
\\

\midrule

& \textit{vanilla}
& \OverallVanilla{16.60}
& \textbf{28.35}
& 9.39
& 15.32
& 24.60
& 9.27
& \textbf{13.57}
& 20.09
& 12.64
& 4.04
& 28.70
\\

\rowcolor{OursBlue}
\ModelCell{\ClaudeIcon}{Claude Code}
& {+\our}
& \OverallOurs{18.74}{2.14}
& 28.15\scoreloss{0.20}
& \textbf{9.48}\scoregain{0.09}
& \textbf{20.30}\scoregain{4.98}
& \textbf{24.75}\scoregain{0.15}
& \textbf{11.06}\scoregain{1.79}
& 12.79\scoreloss{0.78}
& \textbf{21.30}\scoregain{1.21}
& \textbf{17.91}\scoregain{5.27}
& \textbf{7.67}\scoregain{3.63}
& \textbf{33.95}\scoregain{5.25}
\\

\cmidrule(lr){1-13}

& \textit{vanilla}
& \OverallVanilla{17.25}
& 24.26
& 8.90
& 22.39
& 18.93
& 9.29
& 15.70
& \textbf{19.07}
& 15.08
& 5.11
& \textbf{33.71}
\\

\rowcolor{OursBlue}
\ModelCell{\OpenClawIcon}{OpenClaw}
& {+\our}
& \OverallOurs{20.36}{3.11}
& \textbf{30.31}\scoregain{6.05}
& \textbf{12.59}\scoregain{3.68}
& \textbf{22.85}\scoregain{0.45}
& \textbf{25.24}\scoregain{6.32}
& \textbf{16.20}\scoregain{6.90}
& \textbf{17.18}\scoregain{1.47}
& 17.95\scoreloss{1.13}
& \textbf{23.12}\scoregain{8.03}
& \textbf{8.56}\scoregain{3.45}
& 29.62\scoreloss{4.09}
\\

\cmidrule(lr){1-13}

& \textit{vanilla}
& \OverallVanilla{14.49}
& 21.12
& 9.59
& 15.20
& 16.96
& 10.68
& 9.29
& \textbf{20.28}
& 12.05
& 7.05
& 22.73
\\

\rowcolor{OursBlue}
\ModelCell{\OpenScienceIcon}{OpenScience}
& {+\our}
& \OverallOurs{18.09}{3.60}
& \textbf{26.58}\scoregain{5.46}
& \textbf{12.04}\scoregain{2.45}
& \textbf{19.34}\scoregain{4.14}
& \textbf{22.94}\scoregain{5.98}
& \textbf{13.82}\scoregain{3.14}
& \textbf{17.53}\scoregain{8.24}
& 16.02\scoreloss{4.26}
& \textbf{17.45}\scoregain{5.40}
& \textbf{7.06}\scoregain{0.01}
& \textbf{28.11}\scoregain{5.38}
\\

\bottomrule
\end{tabular}%
}

\endgroup

\caption{
Main results on ResearchClawBench.
Panel~A evaluates generalization across backbone LLMs under the fixed
Codex harness, while Panel~B evaluates generalization across agent
harnesses using the fixed DeepSeek-V4-Flash backbone.
\textit{Vanilla} denotes the unmodified configuration, and
{+\our} denotes the same configuration augmented with our
plugin layer.
Bold values indicate the better result within each paired comparison.
Green and red arrows denote absolute improvements and degradations,
respectively.
}

\label{tab:main_results}
\end{table*}

\subsection{Experimental Setup}
\label{sec:experimental_setup}

\paragraph{Benchmarks and Metrics.}
We evaluate {\our} on ResearchClawBench~\cite{ResearchClawBench} and the End-to-End Discovery category of AstaBench~\cite{AstaBench}. 
ResearchClawBench contains 40 tasks across ten scientific domains.
AstaBench evaluates end-to-end AI and NLP research from experimentation to report generation, and we randomly sample 20 tasks from its Easy split.
We report the average rubric-based score on both benchmarks and task completion on AstaBench. Further details are provided in Appendix~A.

\paragraph{Implementation Details.}
We compare each vanilla agent with its AutoSciRub-augmented counterpart under identical task inputs, tools, and execution environments. For ResearchClawBench, we evaluate both cross-model and cross-harness generalization. In cross-model evaluation, we fix Codex~\cite{openai2026codex} as the agent harness and use GPT-5.4~\cite{openai2026gpt54}, GLM-5.2~\cite{zai2026glm52}, and MiniMax-M3~\cite{minimax2026m3} as backbone models. In cross-harness evaluation, we fix DeepSeek-V4-Flash~\cite{deepseek_v4} as the backbone model and compare Claude Code~\cite{anthropic2026claudecode}, OpenClaw~\cite{openclaw2026}, and OpenScience~\cite{syntheticsciences2026openscience}. Following the benchmark-specific evaluation protocols, we use GPT-5.1~\cite{openai2025gpt51} to evaluate report quality.

For AstaBench, we evaluate three representative agent configurations:
Claude Code with DeepSeek-V4-Flash, OpenClaw with DeepSeek-V4-Flash, and Codex with GPT-5.4-mini. The generated artifacts are evaluated using MiniMax-M3 as the evaluator. Across both benchmarks, each resulting submission is independently scored three times, and the mean score is reported.
Additional implementation details, including agent configurations, prompts, and execution settings, are provided in Appendix A.

\begin{figure*}[!t]
  \centering

  \begin{subfigure}[!t]{0.23\textwidth}
      \centering
      \includegraphics[width=\linewidth]{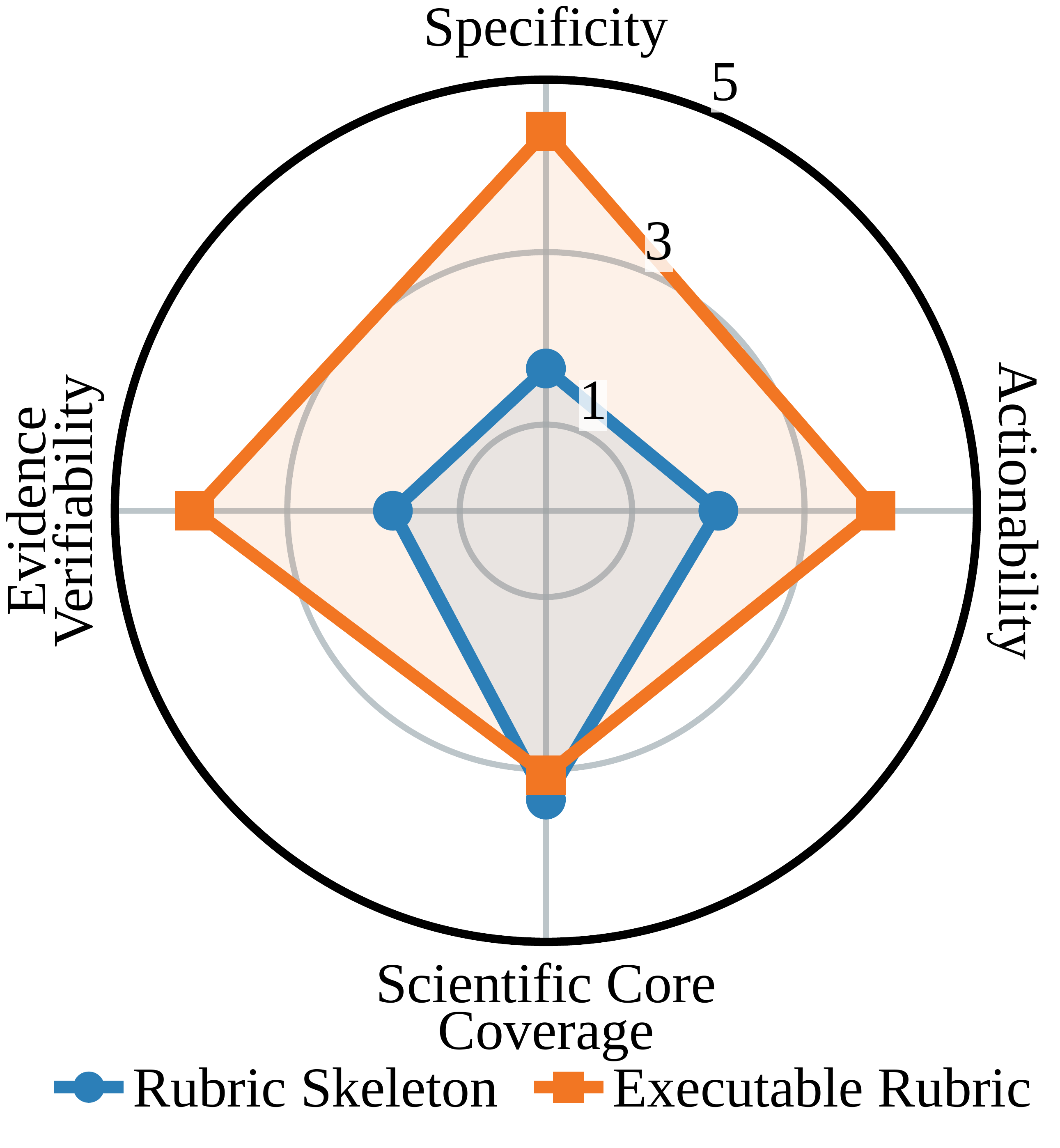}
      \caption{Rubric quality}
      \label{fig:rubric_radar}
  \end{subfigure}
  \hfill
  \begin{subfigure}[!t]{0.34\textwidth}
      \centering
      \includegraphics[width=\linewidth]{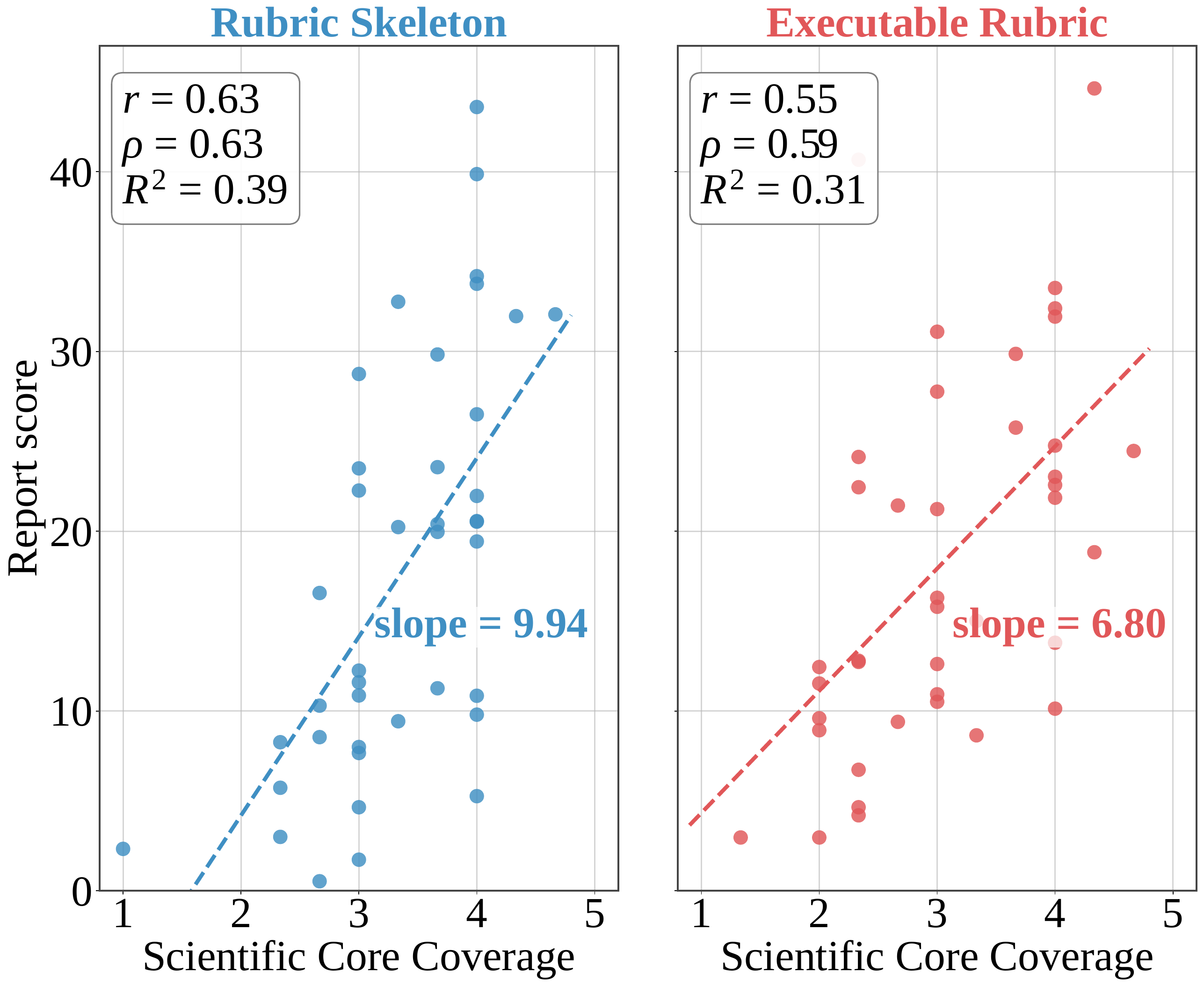}
      \caption{Scientific core coverage vs.\ report score.}
      \label{fig:rubric_core_coverage}
  \end{subfigure}
  \hfill
  \begin{subfigure}[!t]{0.42\textwidth}
      \centering
      \includegraphics[width=\linewidth]{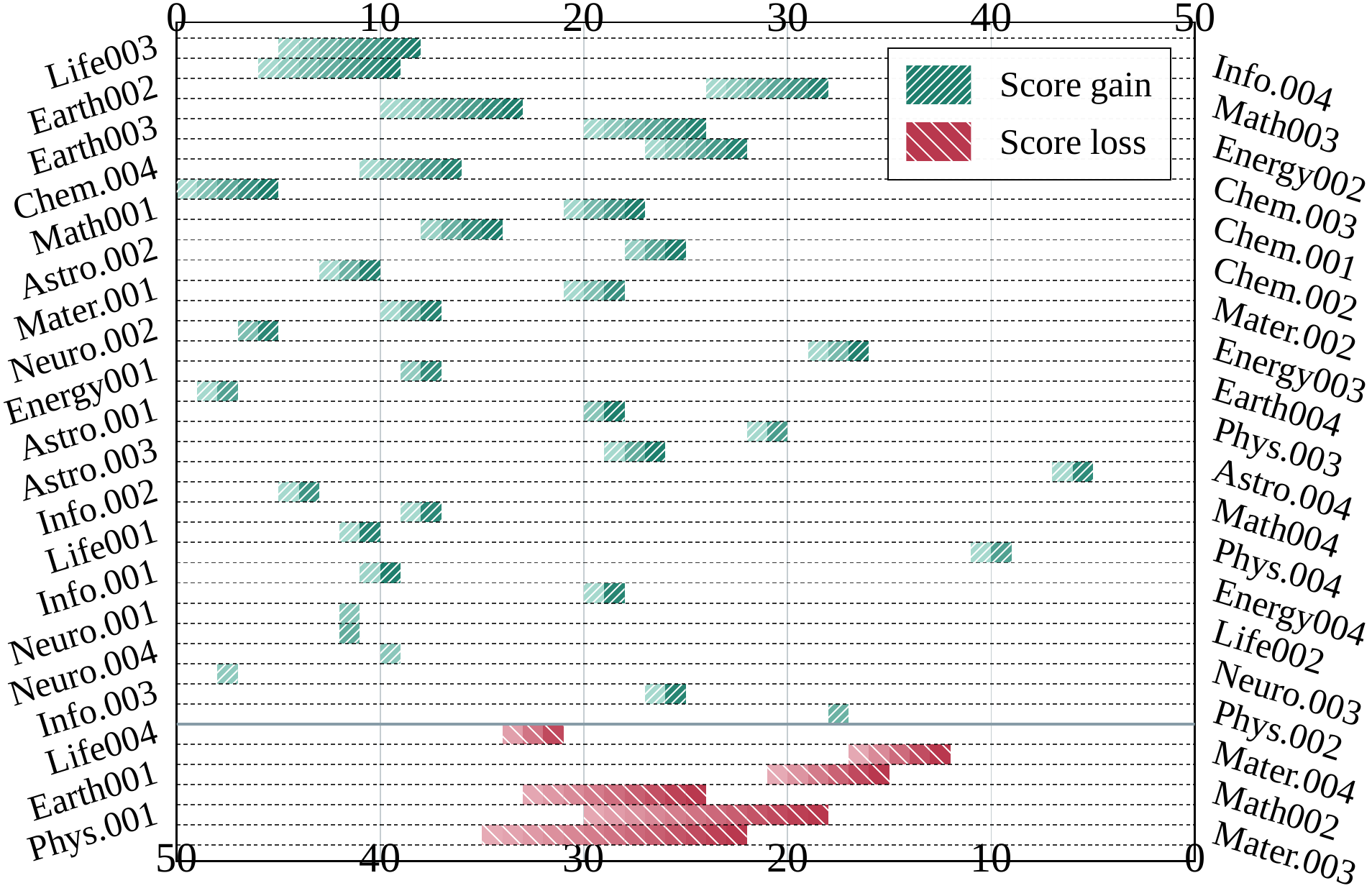}
      \caption{Task-level score changes.}
      \label{fig:rubric_score_changes}
  \end{subfigure}
  \vspace{-1mm}
  \caption{
  Rubric quality and its downstream effects.
  (a) Mean quality profiles of rubric skeletons and executable rubrics
  across 40 tasks.
  (b)  Association between scientific core coverage and downstream
  report score.
  (c) Task-level report-score changes from rubric skeletons to executable
  rubrics.
  }
  \vspace{-2mm}
  \label{fig:rubric_analysis}
\end{figure*}

\begin{figure}[!t]
  \centering
  \includegraphics[width=\linewidth]{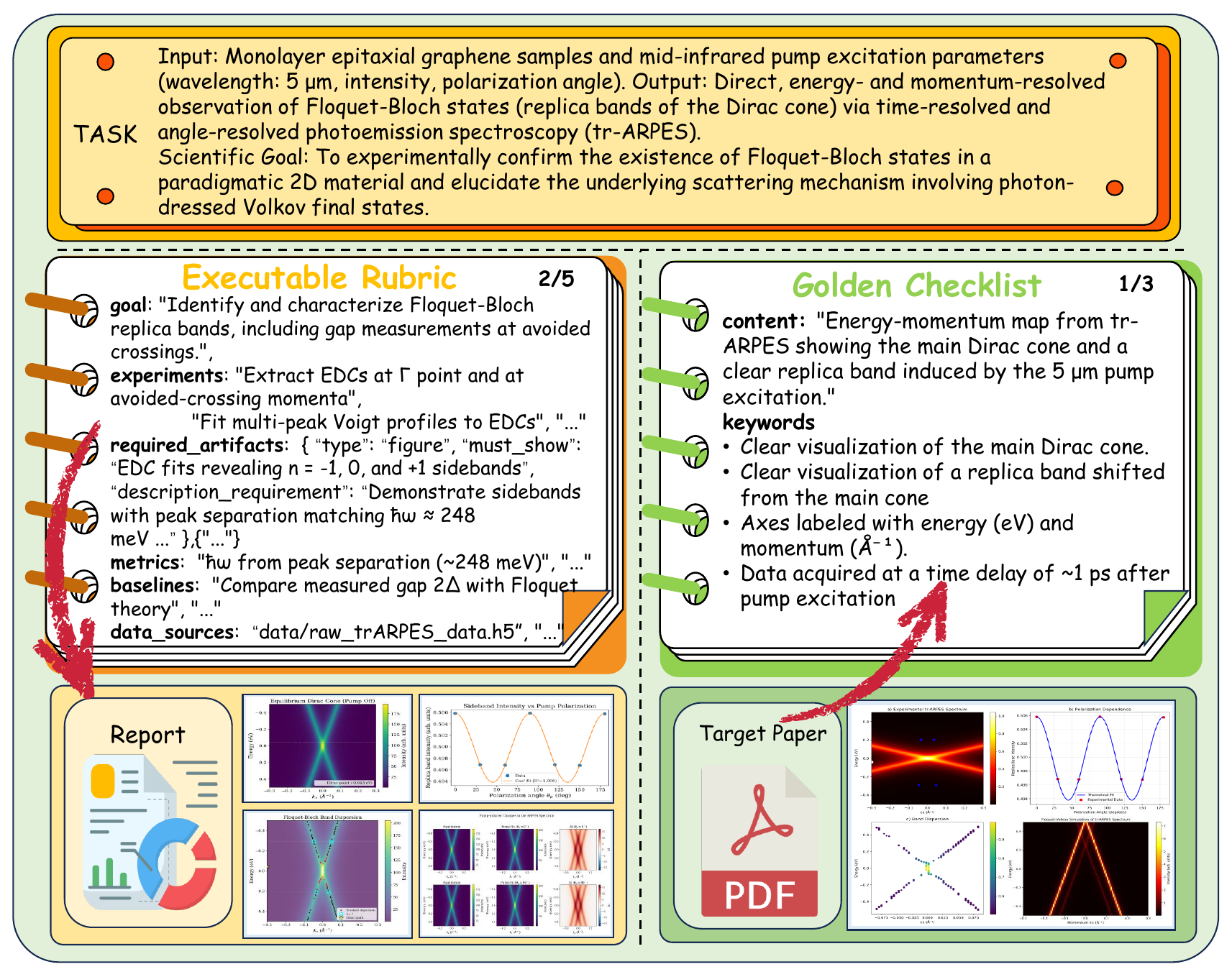}
  \caption{
  Comparison between an AutoSciRub executable rubric and the
  benchmark-provided golden checklist.
  }
  \vspace{-2mm}
  \label{fig:rubric_cmp}
\end{figure}

\subsection{Overall Performance}
\label{sec:overall_performance}

\paragraph{Results on ResearchClawBench.}
As shown in Table~\ref{tab:main_results}, {\our} improves the overall ResearchClawBench score in all six model--harness configurations.
Under the fixed Codex harness, {\our} yields gains of \textbf{2.38} points for GPT-5.4, \textbf{1.87} points for GLM-5.2, and \textbf{1.99} points for MiniMax-M3.
These gains remain consistent across model families and baseline strengths.
Among these configurations, GLM-5.2 with {\our} achieves the highest overall score of \textbf{22.73}.
Together, the consistent gains across three backbone LLMs indicate that the effectiveness of {\our} is not model-specific.

The improvements are similarly preserved when varying the agent harness.
With DeepSeek-V4-Flash as the shared backbone, {\our} yields gains of \textbf{2.14} points for Claude Code, \textbf{3.11} points for OpenClaw, and \textbf{3.60} points for OpenScience.
The larger OpenClaw and OpenScience gains further show that {\our} remains effective across agent systems with distinct workflows and baseline capabilities.
At the domain level, {\our} improves 49 of 60 paired comparisons, including consistent gains in chemistry, energy science, and neuroscience across all six configurations.
Overall, the improvements across backbone models, agent harnesses, and scientific domains demonstrate that {\our} generalizes well to heterogeneous scientific research agents.

\paragraph{Generalization to AstaBench.}
We further evaluate whether these benefits transfer to a different task collection and evaluation protocol.
As shown in Figure~\ref{fig:astabench_results}, {\our} substantially improves the mean scores of all three evaluated agents on the fixed 20-task AstaBench End-to-End Discovery subset.
It yields gains of \textbf{19.36} points for Claude Code, \textbf{12.61} points for Codex, and \textbf{18.38} points for OpenClaw, corresponding to an average improvement of \textbf{16.78} points.
Claude Code and Codex also increase their successful-task counts from 18/20 to 20/20.
OpenClaw completes 19/20 tasks under both settings while gaining \textbf{18.38} points, showing that {\our} improves the quality of completed research outputs in addition to task completion.

\begin{figure*}[!t]
\centering

\begin{subfigure}[!t]{0.34\textwidth}
    \centering
    \includegraphics[width=\linewidth]{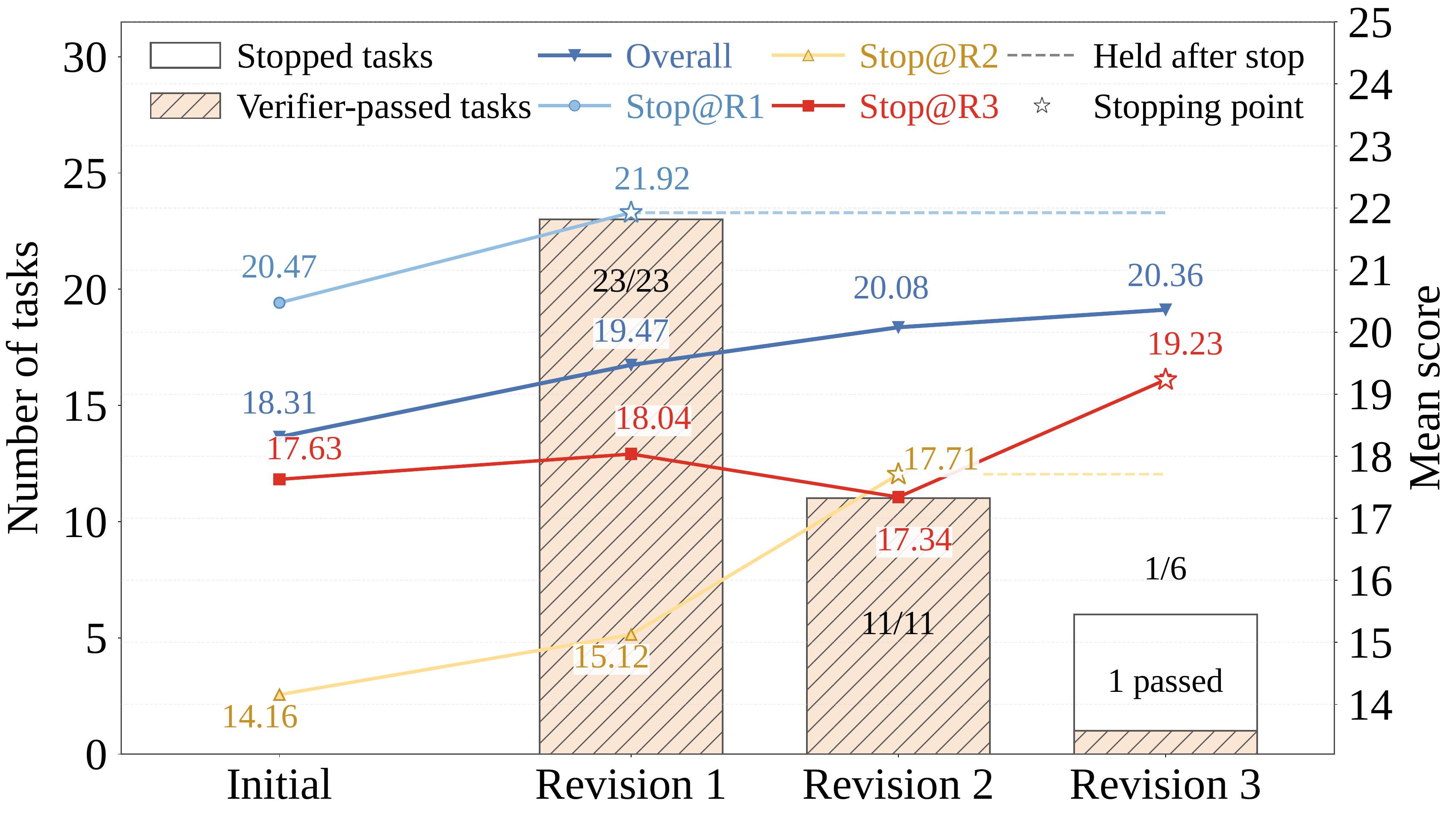}
    \caption{Overall refinement dynamics.}
    \label{fig:refine-overall}
\end{subfigure}
\hfill
\begin{subfigure}[!t]{0.31\textwidth}
    \centering
    \includegraphics[width=\linewidth]{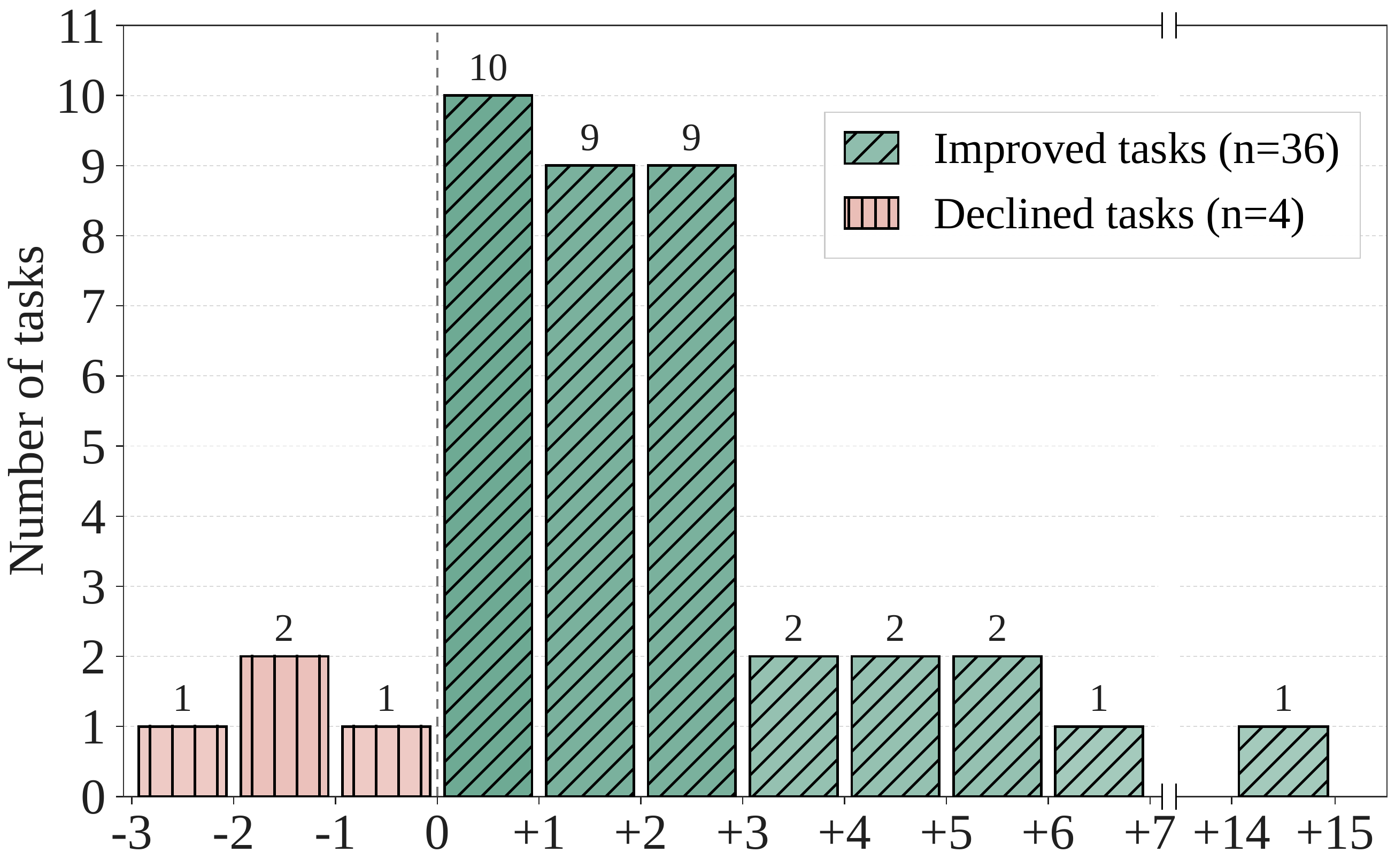}
    \caption{Distribution of task-level score changes.}
    \label{fig:improve-tasks-distribution}
\end{subfigure}
\hfill
\begin{subfigure}[!t]{0.332\textwidth}
    \centering
    \includegraphics[width=\linewidth]{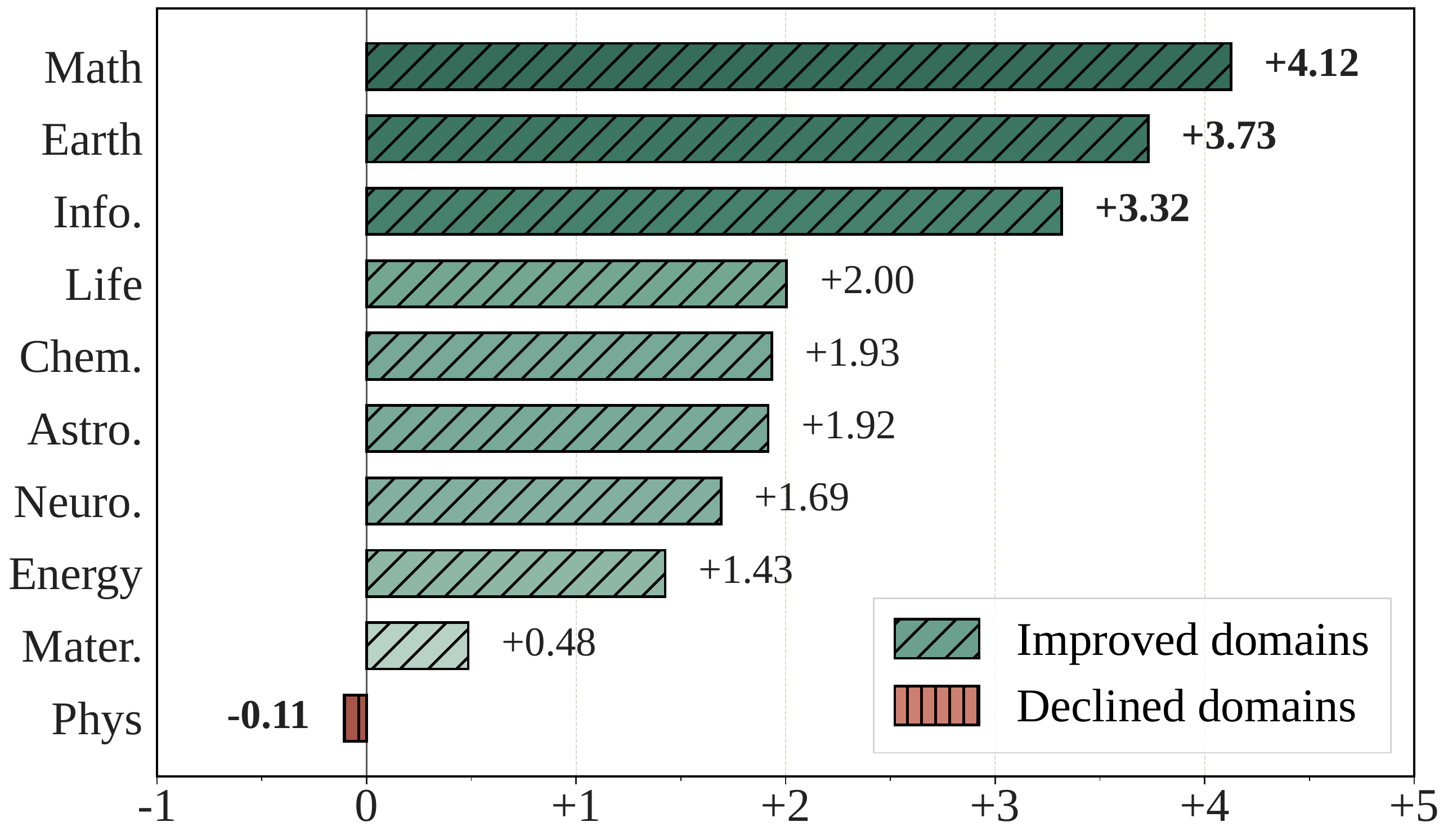}
    \caption{Domain-level mean improvement.}
    \label{fig:domain-improvement}
\end{subfigure}
\vspace{-2mm}
\caption{
    Analysis of rubric-guided refinement.
    (a) Overall score and task stopping behavior across refinement steps.
    (b) Distribution of score changes from the initial reports to the final reports.
    (c) Mean score improvement across scientific domains.
}
\vspace{-2mm}
\label{fig:refinement-analysis}
\end{figure*}

\subsection{Ablation Study}
\label{sec:ablation}

\begin{table}[t]
  \centering

  \small
  \setlength{\tabcolsep}{4.5pt}
  \renewcommand{\arraystretch}{1.15}

  \begin{tabular}{lccccc}
  \toprule
  Setting
  & \textsc{Skel.}
  & \textsc{Grd.}
  & \textsc{Rev.}
  & Score$\uparrow$
  & $\Delta_{\mathrm{Base}}$
  \\
  \midrule

  Base
  & \xmark
  & \xmark
  & \xmark
  & 17.25
  & --
  \\

  w/ Skeleton
  & \cmark
  & \xmark
  & \xmark
  & 17.61
  & +0.36
  \\

  w/ Grounded Rubric
  & \cmark
  & \cmark
  & \xmark
  & 18.31
  & +1.06
  \\

  Full
  & \cmark
  & \cmark
  & \cmark
  & \textbf{20.36}
  & \textbf{+3.11}
  \\

  \bottomrule
  \end{tabular}

  \caption{
  Stage-wise cumulative ablation on ResearchClawBench.
  \textsc{Skel.} denotes Rubric Skeleton Induction;
  \textsc{Grd.} adds Scientific Literature Grounding,
  Task-Data Exploration, and Criterion Synthesis;
  and \textsc{Rev.} adds Rubric-Guided Iterative Revision.
  Scores are averaged over the same 40 tasks, and
  $\Delta_{\mathrm{Base}}$ denotes the improvement over the base setting.
  }

  \label{tab:ablation}
\end{table}

Table~\ref{tab:ablation} presents a cumulative stage-wise ablation on all 40 ResearchClawBench tasks using OpenClaw with DeepSeek-V4-Flash. Starting from the vanilla agent, Rubric Skeleton Induction (\textsc{Skel.}) increases the score from 17.25 to 17.61. This modest gain suggests that making the instruction's implicit scientific requirements explicit helps the agent identify what should be investigated, but offers limited guidance on how these requirements should be carried out and verified.

Adding the grounded-rubric stage (\textsc{Grd.}) further raises the score to 18.31, corresponding to a gain of 1.06 points over the base and an additional 0.70 points over skeleton induction alone. By grounding the induced goals in scientific literature, examining the task-visible data, and synthesizing feasible experiments and evidence requirements, this stage turns a high-level rubric skeleton into more concrete and verifiable criteria.

Rubric-Guided Iterative Revision (\textsc{Rev.}) achieves the best score of 20.36, improving by 3.11  over the base and 2.05  over the grounded-rubric setting. This largest incremental gain indicates that the rubric is most effective when it is used not only to guide execution, but also to identify unmet criteria and support targeted revision of the report and its supporting artifacts. Overall, the monotonic gains show that rubric skeleton induction, grounding, and iterative revision play complementary roles in turning underspecified research instructions into better-supported research outputs.

\subsection{Analysis}
\label{sec:analysis_rubric}

\paragraph{Rubric Induction Produces Executable Guidance.}
Open-ended research instructions rarely specify the experiments, evidence, and success conditions needed to complete a task.
Figure~\ref{fig:rubric_cmp} shows how AutoSciRub addresses this underspecification. While the benchmark checklist mainly describes what should appear in the final report, the executable rubric also specifies how the required evidence should be produced and verified through concrete experiments, artifacts, metrics, baselines, and data sources. It therefore serves as an execution-time specification rather than only a post-hoc evaluation instrument.

For each of the 40 ResearchClawBench tasks, we compare the instruction-derived rubric skeleton with the task-specific executable rubric produced after literature grounding, task-data exploration, and criterion synthesis. As shown in Figure~\ref{fig:rubric_radar}, the mean score across the four evaluated dimensions increases from  2.20 to 3.84. The largest improvement occurs in specificity, which
increases from 1.65 to 4.40 (\(+2.75\)), followed by evidence verifiability from 1.78 to 4.08 (\(+2.30\)) and actionability from 2.00 to 3.83 (\(+1.83\)). These improvements show that rubric induction effectively operationalizes broad scientific goals into concrete, actionable, and verifiable requirements for research execution and verification.

\paragraph{Rubric Induction Improves Operationalization but Not Scientific Framing.}
Although rubric induction substantially improves specificity, actionability, and evidence verifiability, scientific core coverage slightly decreases from 3.35 to 3.07. This contrast suggests that AutoSciRub is more effective at translating an identified scientific direction into concrete and verifiable requirements than at revising the underlying framing of the research problem.

As shown in Figure~\ref{fig:rubric_core_coverage}, scientific core coverage remains positively associated with report quality for both rubric skeletons and executable rubrics. Figure~\ref{fig:rubric_score_changes} further shows that executable rubrics yield a mean report-score gain of 0.70 across all 40 tasks, with improvements observed on 34 tasks. These results suggest that AutoSciRub is effective at translating an identified scientific core into concrete experiments and verifiable evidence requirements, but is less effective at identifying that core when it is missing from the initial rubric skeleton. In such cases, grounding and synthesis may elaborate secondary or off-target analyses rather than correct the research direction. Higher-level scientific judgment therefore remains largely dependent on the backbone model.





\begin{table}[t]
\centering
\setlength{\tabcolsep}{2pt}
\renewcommand{\arraystretch}{1.05}

\resizebox{\columnwidth}{!}{%
\begin{tabular}{
@{}
c|
*{2}{>{\centering\arraybackslash}m{2.25cm}}|
*{2}{>{\centering\arraybackslash}m{2.25cm}}
@{}
}
\toprule

\multirow{2}{*}{Revision Round}
& \multicolumn{2}{c|}{Rubric-Free Self-Refinement}
& \multicolumn{2}{c}{Rubric-Guided Revision} \\

\cmidrule(lr){2-3}
\cmidrule(lr){4-5}

& Score & $\Delta$
& Score & $\Delta$ \\

\midrule

1
& 18.80 & +0.49
& 19.47 & +1.16 \\

2
& 18.52 & +0.21
& 20.08 & +1.77 \\

3
& 19.08 & +0.77
& \textbf{20.36} & \textbf{+2.05} \\

\bottomrule
\end{tabular}%
}

\caption{
Comparison of rubric-free self-refinement and rubric-guided revision across three revision rounds on all 40 ResearchClawBench tasks.
$\Delta$ values denote  score changes relative to the shared unrevised baseline score of 18.31.
}
\label{tab:self_refine_comparison}
\end{table}
\paragraph{Rubric-Guided Revision Outperforms Generic Self-Refinement.}
We analyze whether the improvement from iterative revision is caused by rubric guidance or merely by giving the agent additional opportunities to rewrite its report. Starting from the same checkpoint-0 reports, we compare rubric-guided revision with a rubric-free self-refinement baseline, which asks the agent to inspect and improve its previous output without criterion-level feedback. As shown in Table~\ref{tab:self_refine_comparison}, rubric-free refinement increases the average score from 18.31 to only 19.08 after three rounds and even causes a regression at checkpoint 2. In contrast, rubric-guided revision improves the score monotonically to 20.36, yielding a cumulative improvement of \(2.05\), compared with only \(0.77\) without rubrics. Thus, rubric-guided revision achieves approximately \(2.7\) times the cumulative improvement of rubric-free self-refinement. This result shows that the gain does not come from repeated rewriting alone: explicit criteria identify concrete deficiencies and direct the agent toward unsatisfied scientific requirements, whereas generic self-refinement produces less stable improvements.

Figure~\ref{fig:refine-overall} further shows how the guided improvement accumulates across revision rounds. The first revision raises the average score from 18.31 to 19.47, contributing an improvement of \(1.16\).
The second and third rounds add improvements of \(0.61\) and \(0.28\), respectively. This decrease in aggregate gain mainly results from adaptive stopping rather than ineffective later revisions. After the first round, 23 tasks have already passed the verifier, leaving only 17 tasks for a second revision and six for a third. For these remaining tasks, the average per-round improvements are still \(1.43\) and \(1.89\), respectively. Overall, 35 of the 40 tasks pass within three revisions.
These results suggest that one revision offers a strong default cost/performance trade-off, while additional rounds provide targeted
improvements for harder tasks.

The task- and domain-level results in Figures~\ref{fig:improve-tasks-distribution} and~\ref{fig:domain-improvement} show that the overall gain is broadly distributed rather than dominated by a few high-improvement cases.
Specifically, 36 of the 40 reports improve over their initial versions, and positive average gains are observed in most scientific domains. Mathematics, earth science, and information science achieve the largest average improvements of \(4.12\), \(3.73\), and \(3.32\), respectively. These results demonstrate that the improvement of rubric-guided revision is consistent across tasks and domains, rather than being driven by a single high-scoring case.

\section{Conclusion}

We introduced {\our}, a general framework that automatically induces task-specific scientific rubrics and uses them as execution-time guidance for autonomous research agents. {\our} decomposes underspecified research instructions into scientific goals, grounds evaluation criteria in external evidence, and guides criterion-level verification and revision of generated research artifacts. Experiments on ResearchClawBench and AstaBench show that AutoSciRub consistently improves different backbone models and agent harnesses, demonstrating its effectiveness and generalizability as a plugin layer for autonomous scientific research.

\bibliography{aaai2027}

\begin{thebibliography}{56}
\providecommand{\natexlab}[1]{#1}

\bibitem[{{Anthropic}(2026)}]{anthropic2026claudecode}
{Anthropic}. 2026.
\newblock {Claude Code}.
\newblock \url{https://github.com/anthropics/claude-code}.

\bibitem[{Bragg et~al.(2025)Bragg, D'Arcy, Balepur, Bareket, Dalvi et~al.}]{AstaBench}
Bragg, J.; D'Arcy, M.; Balepur, N.; Bareket, D.; Dalvi, B.; et~al. 2025.
\newblock AstaBench: Rigorous Benchmarking of {AI} Agents with a Scientific Research Suite.
\newblock \emph{CoRR}, abs/2510.21652.

\bibitem[{Chan et~al.(2025)Chan, Chowdhury, Jaffe, Aung, Sherburn et~al.}]{MLE-bench}
Chan, J.~S.; Chowdhury, N.; Jaffe, O.; Aung, J.; Sherburn, D.; et~al. 2025.
\newblock MLE-bench: Evaluating Machine Learning Agents on Machine Learning Engineering.
\newblock In \emph{The Thirteenth International Conference on Learning Representations, {ICLR} 2025, Singapore, April 24-28, 2025}. OpenReview.net.

\bibitem[{Chen et~al.(2026{\natexlab{a}})Chen, Han, Yan, Zhu, Sun, and Che}]{rubric_survey}
Chen, H.; Han, Z.; Yan, Y.; Zhu, Q.; Sun, M.; and Che, W. 2026{\natexlab{a}}.
\newblock From Holistic Evaluation to Structured Criteria: Rubrics Across the Evolving {LLM} Landscape.
\newblock \emph{CoRR}, abs/2606.08625.

\bibitem[{Chen et~al.(2026{\natexlab{b}})Chen, Maiga, Rahmani, and Yilmaz}]{chen2026automated}
Chen, Y.; Maiga, A.; Rahmani, H.~A.; and Yilmaz, E. 2026{\natexlab{b}}.
\newblock Automated Rubrics for Reliable Evaluation of Medical Dialogue Systems.
\newblock \emph{arXiv preprint arXiv:2601.15161}.

\bibitem[{Chen et~al.(2026{\natexlab{c}})Chen, Xiao, Zhao, Xia, Xu, Fang, Li, Zheng, Wang, Xue, Zhang, Li, and Zhang}]{lightmemego}
Chen, Y.; Xiao, B.; Zhao, Y.; Xia, H.; Xu, B.; Fang, J.; Li, Y.; Zheng, Y.; Wang, X.; Xue, Z.; Zhang, L.; Li, H.; and Zhang, N. 2026{\natexlab{c}}.
\newblock LightMem-Ego: Your {AI} Memory for Everyday Life.
\newblock \emph{CoRR}, abs/2607.11487.

\bibitem[{Chen et~al.(2025)Chen, Chen, Ning, Zhang, Wang et~al.}]{ScienceAgentBench}
Chen, Z.; Chen, S.; Ning, Y.; Zhang, Q.; Wang, B.; et~al. 2025.
\newblock ScienceAgentBench: Toward Rigorous Assessment of Language Agents for Data-Driven Scientific Discovery.
\newblock In \emph{The Thirteenth International Conference on Learning Representations, {ICLR} 2025, Singapore, April 24-28, 2025}. OpenReview.net.

\bibitem[{DeepSeek{-}AI(2026)}]{deepseek_v4}
DeepSeek{-}AI. 2026.
\newblock DeepSeek-V4: Towards Highly Efficient Million-Token Context Intelligence.
\newblock \emph{CoRR}, abs/2606.19348.

\bibitem[{Dhole and Agichtein(2026)}]{RubricRAG}
Dhole, K.~D.; and Agichtein, E. 2026.
\newblock {RubricRAG}: Towards Interpretable and Reliable {LLM} Evaluation via Domain Knowledge Retrieval for Rubric Generation.
\newblock In Moffat, A.; Scholer, F.; Bast, H.; Najork, M.; and Zhang, M., eds., \emph{Proceedings of the 49th International {ACM} {SIGIR} Conference on Research and Development in Information Retrieval, {SIGIR} 2026, Melbourne, VIC, Australia, July 20--24, 2026}, 3681--3687. {ACM}.

\bibitem[{Ding(2026)}]{AdaRubric}
Ding, L. 2026.
\newblock AdaRubric: Task-Adaptive Rubrics for {LLM} Agent Evaluation.
\newblock \emph{CoRR}, abs/2603.21362.

\bibitem[{Fan et~al.(2024)Fan, Wang, Wu, and Zhang}]{SedarEval}
Fan, Z.; Wang, W.; Wu, X.; and Zhang, D. 2024.
\newblock SedarEval: Automated Evaluation using Self-Adaptive Rubrics.
\newblock In Al{-}Onaizan, Y.; Bansal, M.; and Chen, Y., eds., \emph{Findings of the Association for Computational Linguistics: {EMNLP} 2024, Miami, Florida, USA, November 12-16, 2024}, volume {EMNLP} 2024 of \emph{Findings of {ACL}}, 16916--16930. Association for Computational Linguistics.

\bibitem[{Gao, Fang, and Zitnik(2026)}]{AutoScientists}
Gao, S.; Fang, A.; and Zitnik, M. 2026.
\newblock AutoScientists: Self-Organizing Agent Teams for Long-Running Scientific Experimentation.
\newblock \emph{CoRR}, abs/2605.28655.

\bibitem[{Gao et~al.(2026)Gao, Su, Sui, Ginder, and Zitnik}]{Qworld}
Gao, S.; Su, Y.; Sui, P.; Ginder, C.; and Zitnik, M. 2026.
\newblock Qworld: Question-Specific Evaluation Criteria for LLMs.
\newblock \emph{CoRR}, abs/2603.23522.

\bibitem[{Garikaparthi, Patwardhan, and Cohan(2026)}]{ResearchGym}
Garikaparthi, A.; Patwardhan, M.; and Cohan, A. 2026.
\newblock ResearchGym: Evaluating Language Model Agents on Real-World {AI} Research.
\newblock \emph{CoRR}, abs/2602.15112.

\bibitem[{Ghareeb et~al.(2026)Ghareeb, Chang, Mitchener, Yiu, Szostkiewicz et~al.}]{Robin}
Ghareeb, A.~E.; Chang, B.; Mitchener, L.; Yiu, A.; Szostkiewicz, C.~J.; et~al. 2026.
\newblock A Multi-Agent System for Automating Scientific Discovery.
\newblock \emph{Nature}, 655: 497--505.

\bibitem[{Gottweis et~al.(2026)Gottweis, Weng, Daryin, Tu, Sirkovic et~al.}]{accelerating}
Gottweis, J.; Weng, W.-H.; Daryin, A.; Tu, T.; Sirkovic, P.; et~al. 2026.
\newblock Accelerating Scientific Discovery with {Co-Scientist}.
\newblock \emph{Nature}, 655: 487--496.

\bibitem[{Gou et~al.(2024)Gou, Shao, Gong, Shen, Yang, Duan, and Chen}]{CRITIC}
Gou, Z.; Shao, Z.; Gong, Y.; Shen, Y.; Yang, Y.; Duan, N.; and Chen, W. 2024.
\newblock {CRITIC:} Large Language Models Can Self-Correct with Tool-Interactive Critiquing.
\newblock In \emph{The Twelfth International Conference on Learning Representations, {ICLR} 2024, Vienna, Austria, May 7-11, 2024}. OpenReview.net.

\bibitem[{Hashemi et~al.(2024)Hashemi, Eisner, Rosset, Durme, and Kedzie}]{LLM_Rubric}
Hashemi, H.; Eisner, J.; Rosset, C.; Durme, B.~V.; and Kedzie, C. 2024.
\newblock LLM-Rubric: {A} Multidimensional, Calibrated Approach to Automated Evaluation of Natural Language Texts.
\newblock In Ku, L.; Martins, A.; and Srikumar, V., eds., \emph{Proceedings of the 62nd Annual Meeting of the Association for Computational Linguistics (Volume 1: Long Papers), {ACL} 2024, Bangkok, Thailand, August 11-16, 2024}, 13806--13834. Association for Computational Linguistics.

\bibitem[{Hong et~al.(2026)Hong, Li, Chen, Huy, Ananiadou, Kim, and Lin}]{hong2026can}
Hong, H.; Li, Y.; Chen, J.; Huy, L.~G.; Ananiadou, S.; Kim, J.-j.; and Lin, C. 2026.
\newblock Can LLMs Write Reliable Rubrics? A Meta-Evaluation for Experiment Reproduction.
\newblock \emph{arXiv preprint arXiv:2607.12835}.

\bibitem[{Ifargan et~al.(2024)Ifargan, Hafner, Kern, Alcalay, and Kishony}]{data2paper}
Ifargan, T.; Hafner, L.; Kern, M.; Alcalay, O.; and Kishony, R. 2024.
\newblock Autonomous LLM-driven research from data to human-verifiable research papers.
\newblock \emph{CoRR}, abs/2404.17605.

\bibitem[{LeVine et~al.(2026)LeVine, Evers, Saltwick, and Venkatesh}]{RubricRefine}
LeVine, W.; Evers, B.; Saltwick, S.; and Venkatesh, A. 2026.
\newblock RubricRefine: Improving Tool-Use Agent Reliability with Training-Free Pre-Execution Refinement.
\newblock \emph{CoRR}, abs/2605.09730.

\bibitem[{Li et~al.(2026{\natexlab{a}})Li, Liu, Wang, Huang, Li, Jia, Hu, and Zhang}]{lycheememoryv2}
Li, D.; Liu, Z.; Wang, J.; Huang, J.; Li, F.; Jia, B.; Hu, B.; and Zhang, M. 2026{\natexlab{a}}.
\newblock LycheeMemory V2: Efficient Long-Term Memory for LLM Agents via Semantic Segment-Level Consolidation.
\newblock arXiv:2608.12990.

\bibitem[{Li et~al.(2026{\natexlab{b}})Li, Du, Xu, Zhu, Wang et~al.}]{DeepResearch_Bench2}
Li, R.; Du, M.; Xu, B.; Zhu, C.; Wang, X.; et~al. 2026{\natexlab{b}}.
\newblock DeepResearch Bench {II:} Diagnosing Deep Research Agents via Rubrics from Expert Report.
\newblock \emph{CoRR}, abs/2601.08536.

\bibitem[{Liu et~al.(2026)Liu, Qiu, Li, Li, Ji et~al.}]{AutoResearchClaw}
Liu, J.; Qiu, S.; Li, M.; Li, B.; Ji, H.; et~al. 2026.
\newblock AutoResearchClaw: Self-Reinforcing Autonomous Research with Human-AI Collaboration.
\newblock \emph{CoRR}, abs/2605.20025.

\bibitem[{Lu et~al.(2024)Lu, Lu, Lange, Foerster, Clune et~al.}]{ai_scientist}
Lu, C.; Lu, C.; Lange, R.~T.; Foerster, J.~N.; Clune, J.; et~al. 2024.
\newblock The {AI} Scientist: Towards Fully Automated Open-Ended Scientific Discovery.
\newblock \emph{CoRR}, abs/2408.06292.

\bibitem[{Lu et~al.(2026)Lu, Li, Shi, Wang, Wang, and Hu}]{Structured_Episodic_Event_Memory}
Lu, Z.; Li, D.; Shi, Y.; Wang, B.; Wang, L.; and Hu, B. 2026.
\newblock Structured Episodic Event Memory.
\newblock In Liakata, M.; Moreira, V.~P.; Zhang, J.; and Jurgens, D., eds., \emph{Proceedings of the 64th Annual Meeting of the Association for Computational Linguistics (Volume 1: Long Papers), {ACL} 2026, San Diego, California, United States, July 2-7, 2026}, 6125--6141. Association for Computational Linguistics.

\bibitem[{Luo et~al.(2026)Luo, Yu, Wang, Zhu, Zhang, Wei, Du, Zheng, and Chen}]{xKG}
Luo, Y.; Yu, Z.; Wang, X.; Zhu, Y.; Zhang, N.; Wei, L.; Du, L.; Zheng, D.; and Chen, H. 2026.
\newblock What Makes {AI} Research Replicable? Executable Knowledge Graphs as Scientific Knowledge Representations.
\newblock In Liakata, M.; Moreira, V.~P.; Zhang, J.; and Jurgens, D., eds., \emph{Proceedings of the 64th Annual Meeting of the Association for Computational Linguistics (Volume 2: Short Papers), {ACL} 2026, San Diego, California, United States, July 2-7, 2026}, 841--861. Association for Computational Linguistics.

\bibitem[{Lyu et~al.(2026)Lyu, Zhang, Yi, Zhao, Guo et~al.}]{EvoScientist}
Lyu, Y.; Zhang, X.; Yi, X.; Zhao, Y.; Guo, S.; et~al. 2026.
\newblock EvoScientist: Towards Multi-Agent Evolving {AI} Scientists for End-to-End Scientific Discovery.
\newblock \emph{CoRR}, abs/2603.08127.

\bibitem[{Madaan et~al.(2023)Madaan, Tandon, Gupta, Hallinan, Gao, Wiegreffe, Alon, Dziri, Prabhumoye, Yang, Gupta, Majumder, Hermann, Welleck, Yazdanbakhsh, and Clark}]{Self-Refine}
Madaan, A.; Tandon, N.; Gupta, P.; Hallinan, S.; Gao, L.; Wiegreffe, S.; Alon, U.; Dziri, N.; Prabhumoye, S.; Yang, Y.; Gupta, S.; Majumder, B.~P.; Hermann, K.; Welleck, S.; Yazdanbakhsh, A.; and Clark, P. 2023.
\newblock Self-Refine: Iterative Refinement with Self-Feedback.
\newblock In Oh, A.; Naumann, T.; Globerson, A.; Saenko, K.; Hardt, M.; and Levine, S., eds., \emph{Advances in Neural Information Processing Systems 36: Annual Conference on Neural Information Processing Systems 2023, NeurIPS 2023, New Orleans, LA, USA, December 10 - 16, 2023}.

\bibitem[{{MiniMax}(2026)}]{minimax2026m3}
{MiniMax}. 2026.
\newblock {MiniMax M3}: Frontier Coding, 1M Context, Native Multimodality---All in One Model.
\newblock \url{https://www.minimax.io/blog/minimax-m3}.
\newblock Accessed: 2026-07-24.

\bibitem[{Mitchener et~al.(2025)Mitchener, Yiu, Chang, Bourdenx, Nadolski et~al.}]{Kosmos}
Mitchener, L.; Yiu, A.; Chang, B.; Bourdenx, M.; Nadolski, T.; et~al. 2025.
\newblock Kosmos: An {AI} Scientist for Autonomous Discovery.
\newblock \emph{CoRR}, abs/2511.02824.

\bibitem[{{OpenAI}(2025)}]{openai2025gpt51}
{OpenAI}. 2025.
\newblock {GPT-5.1}: A Smarter, More Conversational {ChatGPT}.
\newblock \url{https://openai.com/index/gpt-5-1/}.
\newblock Accessed: 2026-07-24.

\bibitem[{{OpenAI}(2026{\natexlab{a}})}]{openai2026codex}
{OpenAI}. 2026{\natexlab{a}}.
\newblock {Codex}.
\newblock \url{https://github.com/openai/codex}.

\bibitem[{{OpenAI}(2026{\natexlab{b}})}]{openai2026gpt54}
{OpenAI}. 2026{\natexlab{b}}.
\newblock Introducing {GPT-5.4}.
\newblock \url{https://openai.com/index/introducing-gpt-5-4/}.
\newblock Accessed: 2026-07-24.

\bibitem[{{OpenClaw Foundation}(2026)}]{openclaw2026}
{OpenClaw Foundation}. 2026.
\newblock {OpenClaw}.
\newblock \url{https://github.com/openclaw/openclaw}.

\bibitem[{Schmidgall et~al.(2025)Schmidgall, Su, Wang, Sun, Wu et~al.}]{agentlaboratory}
Schmidgall, S.; Su, Y.; Wang, Z.; Sun, X.; Wu, J.; et~al. 2025.
\newblock Agent Laboratory: Using {LLM} Agents as Research Assistants.
\newblock In Christodoulopoulos, C.; Chakraborty, T.; Rose, C.; and Peng, V., eds., \emph{Findings of the Association for Computational Linguistics: {EMNLP} 2025, Suzhou, China, November 4-9, 2025}, 5977--6043. Association for Computational Linguistics.

\bibitem[{Sharma et~al.(2025)Sharma, Zhang, Bandi, Wang, Aich, Nghiem, Rabbani, Htet, Jang, Basu, Balwani, Peskoff, Ayestaran, Hendryx, Kenstler, and Liu}]{ResearchRubrics}
Sharma, M.; Zhang, C. B.~C.; Bandi, C.; Wang, C.; Aich, A.; Nghiem, H.; Rabbani, T.; Htet, Y.; Jang, B.; Basu, S.; Balwani, A.; Peskoff, D.; Ayestaran, M.; Hendryx, S.~M.; Kenstler, B.; and Liu, B. 2025.
\newblock ResearchRubrics: {A} Benchmark of Prompts and Rubrics For Evaluating Deep Research Agents.
\newblock \emph{CoRR}, abs/2511.07685.

\bibitem[{Shen et~al.(2026)Shen, Qiu, Whitehouse, Alazraki, Goel et~al.}]{Rethinking_Rubric}
Shen, W.~F.; Qiu, X.; Whitehouse, C.; Alazraki, L.; Goel, S.; et~al. 2026.
\newblock Rethinking Rubric Generation for Improving {LLM} Judge and Reward Modeling for Open-ended Tasks.
\newblock \emph{CoRR}, abs/2602.05125.

\bibitem[{Siro, Aliannejadi, and Aliannejadi(2026)}]{GER_Eval}
Siro, C.; Aliannejadi, P.; and Aliannejadi, M. 2026.
\newblock Learning to Judge: LLMs Designing and Applying Evaluation Rubrics.
\newblock In Demberg, V.; Inui, K.; and Marquez, L., eds., \emph{Findings of the Association for Computational Linguistics: {EACL} 2026, Rabat, Morocco, March 24-29, 2026}, Findings of {ACL}, 6371--6389. Association for Computational Linguistics.

\bibitem[{Starace et~al.(2025)Starace, Jaffe, Sherburn, Aung, Chan et~al.}]{PaperBench}
Starace, G.; Jaffe, O.; Sherburn, D.; Aung, J.; Chan, J.~S.; et~al. 2025.
\newblock PaperBench: Evaluating AI's Ability to Replicate {AI} Research.
\newblock In Singh, A.; Fazel, M.; Hsu, D.; Lacoste{-}Julien, S.; Berkenkamp, F.; Maharaj, T.; Wagstaff, K.; and Zhu, J., eds., \emph{Forty-second International Conference on Machine Learning, {ICML} 2025, Vancouver, BC, Canada, July 13-19, 2025}, volume 267 of \emph{Proceedings of Machine Learning Research}. {PMLR} / OpenReview.net.

\bibitem[{{Synthetic Sciences}(2026)}]{syntheticsciences2026openscience}
{Synthetic Sciences}. 2026.
\newblock {OpenScience}: The Open-Source AI Workbench for Scientific Research.
\newblock \url{https://github.com/synthetic-sciences/openscience}.

\bibitem[{Tang et~al.(2025)Tang, Xia, Li, and Huang}]{AI-Researcher}
Tang, J.; Xia, L.; Li, Z.; and Huang, C. 2025.
\newblock AI-Researcher: Autonomous Scientific Innovation.
\newblock \emph{CoRR}, abs/2505.18705.

\bibitem[{Wan et~al.(2026)Wan, Fang, Li, Huo, Wang, Mi, Yu, and Lyu}]{wan2026inference}
Wan, Y.; Fang, T.; Li, Z.; Huo, Y.; Wang, W.; Mi, H.; Yu, D.; and Lyu, M.~R. 2026.
\newblock Inference-Time Scaling of Verification: Self-Evolving Deep Research Agents via Test-Time Rubric-Guided Verification.
\newblock In Liakata, M.; Moreira, V.~P.; Zhang, J.; and Jurgens, D., eds., \emph{Findings of the Association for Computational Linguistics, {ACL} 2026, San Diego, California, United States, July 2-7, 2026}, 24822--24835. Association for Computational Linguistics.

\bibitem[{Wang et~al.(2026)Wang, Bai, Luo, Su, Sun, Yu, Liu, Zhou, Cardie, Dredze, Xing, and Hu}]{FIREBench}
Wang, Z.; Bai, F.; Luo, Z.; Su, J.; Sun, K.; Yu, X.; Liu, J.; Zhou, K.; Cardie, C.; Dredze, M.; Xing, E.~P.; and Hu, Z. 2026.
\newblock FIRE-Bench: Evaluating Agents on the Rediscovery of Scientific Insights.
\newblock \emph{CoRR}, abs/2602.02905.

\bibitem[{Wang and Blanco(2026)}]{dynamic_rubric}
Wang, Z.; and Blanco, E. 2026.
\newblock Generating and Refining Dynamic Evaluation Rubrics for LLM-as-a-Judge.
\newblock \emph{CoRR}, abs/2605.30568.

\bibitem[{Wei et~al.(2025)Wei, Yang, Zhang, Chen, Zhuang, Gao, Zhou, Wang, Gao, Cao, Qiu, He, Zhang, You, Zheng, Ding, Ouyang, Dong, Cheng, Sun, Bai, and Zhou}]{agentic_science_survey}
Wei, J.; Yang, Y.; Zhang, X.; Chen, Y.; Zhuang, X.; Gao, Z.; Zhou, D.; Wang, G.; Gao, Z.; Cao, J.; Qiu, Z.; He, X.; Zhang, Q.; You, C.; Zheng, S.; Ding, N.; Ouyang, W.; Dong, N.; Cheng, Y.; Sun, S.; Bai, L.; and Zhou, B. 2025.
\newblock From {AI} for Science to Agentic Science: {A} Survey on Autonomous Scientific Discovery.
\newblock \emph{CoRR}, abs/2508.14111.

\bibitem[{Xu et~al.(2026{\natexlab{a}})Xu, Chen, Fang, Zhong, Yao, Zhu, Du, and Deng}]{structmem}
Xu, B.; Chen, Y.; Fang, J.; Zhong, R.; Yao, Y.; Zhu, Y.; Du, L.; and Deng, S. 2026{\natexlab{a}}.
\newblock StructMem: Structured Memory for Long-Horizon Behavior in LLMs.
\newblock In Liakata, M.; Moreira, V.~P.; Zhang, J.; and Jurgens, D., eds., \emph{Proceedings of the 64th Annual Meeting of the Association for Computational Linguistics (Volume 2: Short Papers), {ACL} 2026, San Diego, California, United States, July 2-7, 2026}, 122--146. Association for Computational Linguistics.

\bibitem[{Xu et~al.(2026{\natexlab{b}})Xu, Xue, Chen, Fu, Wu, Huang, Jiang, Fang, Deng, Chen, Yao, Wang, Shang, Yu, and Zhang}]{tokenpilot}
Xu, B.; Xue, Z.; Chen, D.; Fu, C.; Wu, C.; Huang, C.; Jiang, C.; Fang, J.; Deng, X.; Chen, Y.; Yao, Y.; Wang, X.; Shang, J.; Yu, G.; and Zhang, N. 2026{\natexlab{b}}.
\newblock TokenPilot: Cache-Efficient Context Management for {LLM} Agents.
\newblock \emph{CoRR}, abs/2606.17016.

\bibitem[{Xu et~al.(2026{\natexlab{c}})Xu, Li, Ye, Cao, Chen et~al.}]{ResearchClawBench}
Xu, W.; Li, S.; Ye, T.; Cao, Q.; Chen, Y.; et~al. 2026{\natexlab{c}}.
\newblock ResearchClawBench: {A} Benchmark for End-to-End Autonomous Scientific Research.
\newblock \emph{CoRR}, abs/2606.07591.

\bibitem[{Yamada et~al.(2025)Yamada, Lange, Lu, Hu, Lu et~al.}]{ai_scientist_v2}
Yamada, Y.; Lange, R.~T.; Lu, C.; Hu, S.; Lu, C.; et~al. 2025.
\newblock The {AI} Scientist-v2: Workshop-Level Automated Scientific Discovery via Agentic Tree Search.
\newblock \emph{CoRR}, abs/2504.08066.

\bibitem[{Ye et~al.(2026)Ye, Lin, Tang, Luo, Yang, Su, Thapa, Yang, Liu, Li, Gao, Ding, He, Zhang, Sun, Wang, Zhong, Shen, He, Ma, Ermon, Li, Chu, Wang, and Xu}]{Evaluation_driven}
Ye, H.; Lin, H.; Tang, J.; Luo, Y.; Yang, C.; Su, C.; Thapa, R.; Yang, R.; Liu, R.; Li, Z.; Gao, C.; Ding, D.; He, G.; Zhang, M.; Sun, L.; Wang, W.; Zhong, Y.; Shen, Z.; He, D.; Ma, J.; Ermon, S.; Li, T.; Chu, X.; Wang, J.~Z.; and Xu, Y. 2026.
\newblock Evaluation-driven Scaling for Scientific Discovery.
\newblock \emph{CoRR}, abs/2604.19341.

\bibitem[{Yehudai et~al.(2026)Yehudai, Eden, Li, Uziel, Zhao, Bar{-}Haim, Cohan, and Shmueli{-}Scheuer}]{agent_evaluation_survey}
Yehudai, A.; Eden, L.; Li, A.; Uziel, G.; Zhao, Y.; Bar{-}Haim, R.; Cohan, A.; and Shmueli{-}Scheuer, M. 2026.
\newblock A Survey on Evaluation of LLM-based Agents.
\newblock In Liakata, M.; Moreira, V.~P.; Zhang, J.; and Jurgens, D., eds., \emph{Findings of the Association for Computational Linguistics, {ACL} 2026, San Diego, California, United States, July 2-7, 2026}, 26690--26714. Association for Computational Linguistics.

\bibitem[{Yu et~al.(2026)Yu, Xu, Wang, and Yang}]{ThinkWithRubrics}
Yu, J.; Xu, Z.; Wang, J.; and Yang, Y. 2026.
\newblock Think-with-Rubrics: From External Evaluator to Internal Reasoning Guidance.
\newblock \emph{CoRR}, abs/2605.07461.

\bibitem[{{Z.ai}(2026)}]{zai2026glm52}
{Z.ai}. 2026.
\newblock {GLM-5.2}: Built for Long-Horizon Tasks.
\newblock \url{https://z.ai/blog/glm-5.2}.
\newblock Accessed: 2026-07-24.

\bibitem[{Zheng et~al.(2026)Zheng, Zhang, Luo, Mao, Gao, Du, Chen, and Zhang}]{can_we_pre}
Zheng, J.; Zhang, J.; Luo, Y.; Mao, Y.; Gao, Y.; Du, L.; Chen, H.; and Zhang, N. 2026.
\newblock Can We Predict Before Executing Machine Learning Agents?
\newblock In Liakata, M.; Moreira, V.~P.; Zhang, J.; and Jurgens, D., eds., \emph{Proceedings of the 64th Annual Meeting of the Association for Computational Linguistics (Volume 1: Long Papers), {ACL} 2026, San Diego, California, United States, July 2-7, 2026}, 3941--3974. Association for Computational Linguistics.

\bibitem[{Zhu et~al.(2026)Zhu, Wei, Xu, Cheng, Chen, and He}]{DeepRubric}
Zhu, M.; Wei, C.; Xu, J.; Cheng, Y.; Chen, Z.; and He, J. 2026.
\newblock {DEEPRUBRIC:} Evidence-Tree Rubric Supervision for Efficient Reinforcement Learning of Deep Research Agents.
\newblock \emph{CoRR}, abs/2606.17029.

\end{thebibliography}


\appendix

\section{Additional Experimental and Methodological Details}
\label{app:supplementary}

\subsection{Benchmark and Evaluation Details}
\label{app:benchmark_evaluation}

\paragraph{ResearchClawBench.}
ResearchClawBench contains 40 end-to-end scientific research tasks, with four tasks from each of ten domains: astronomy, chemistry, earth science, energy science, information science, life science, materials science, mathematics, neuroscience, and physics. Each task is constructed from a published target paper with a clearly defined scientific question and accessible raw data. The evaluated agent is provided with a task instruction, related literature, raw data, and a workspace in which it can inspect files, implement analyses, execute experiments, generate figures, and write a final scientific report. The hidden target paper and its official evaluation rubric are not accessible to the agent during rubric induction, research execution, verification, or revision.

We use the official ResearchClawBench evaluator without modifying its scoring protocol. Each benchmark rubric consists of weighted criteria associated with concrete scientific artifacts in the hidden target paper. The rubric contains both textual criteria, which assess scientific claims, methods, quantitative results, and explanations, and image criteria, which assess figure-level evidence. GPT-5.1 is used to evaluate the final report and its supporting artifacts against these criteria. The resulting task score ranges from 0 to 100, with a score around 50 intended to represent approximately target-paper-level re-discovery under the benchmark protocol.

For each configuration, the overall score is the arithmetic mean over all 40 tasks. Each domain score is the arithmetic mean over the four tasks belonging to that domain. All reported improvements are absolute differences between paired \textit{vanilla} and \(+{\our}\) scores on the same task set.

\paragraph{AstaBench End-to-End Discovery.}
We additionally evaluate {\our} on a fixed subset of 20 tasks from the
Easy split of the AstaBench End-to-End Discovery
benchmark. The subset is sampled once from the
original 40-task test set using random seed 20260707. The sampling
procedure does not preserve the order of the source dataset; instead,
the sampled order is retained as the execution order. The same fixed
subset and execution order are used for all evaluated configurations.
Table~\ref{tab:astabench_tasks} lists the exact task identifiers and
task names.

\begin{table*}[t]
\centering
\small
\setlength{\tabcolsep}{5pt}
\renewcommand{\arraystretch}{1.06}
\begin{tabular}{
    c
    p{0.22\textwidth}
    p{0.62\textwidth}
}
\toprule
\textbf{Order}
& \textbf{Task ID}
& \textbf{Task name}
\\
\midrule
1
& \texttt{idea-226-simplified}
& simple-knowledge-enhanced-vqa
\\
2
& \texttt{idea-206-simplified}
& selective-entity-quantization
\\
3
& \texttt{idea-131-simplified}
& simple-semantic-backdoor-defense
\\
4
& \texttt{idea-21-simplified}
& text-to-code-prompt-retrieval
\\
5
& \texttt{idea-78-simplified}
& simple-bias-detection-worldmodel
\\
6
& \texttt{idea-337-simplified}
& medical-verbalization-calibration
\\
7
& \texttt{idea-17-simplified}
& topic-focused-memory-summarization
\\
8
& \texttt{idea-86-simplified}
& simple-language-skill-transfer
\\
9
& \texttt{idea-259-simplified}
& simple-relevance-metrics
\\
10
& \texttt{idea-174-simplified}
& simple-position-aware-prompts
\\
11
& \texttt{idea-5-simplified}
& simple-entity-debate
\\
12
& \texttt{idea-159-simplified}
& confidence-based-contamination-detection
\\
13
& \texttt{idea-304-simplified}
& dual-task-adversarial-robustness
\\
14
& \texttt{idea-192-simplified}
& simple-fact-unlearning
\\
15
& \texttt{idea-147-simplified}
& mini-socratic-code-generation
\\
16
& \texttt{idea-301-simplified}
& simplified-task-learning-transfer
\\
17
& \texttt{idea-134-simplified}
& simple-cross-lingual-backdoor
\\
18
& \texttt{idea-278-simplified}
& visual-entity-knowledge-test
\\
19
& \texttt{idea-68-simplified}
& evolutionary-prompt-optimization
\\
20
& \texttt{idea-141-simplified}
& biobert-pet-ner
\\
\bottomrule
\end{tabular}
\caption{
The fixed 20-task subset used for AstaBench evaluation, listed in
the retained execution order. The tasks were sampled from the
original 40-task test set using random seed 20260707.
}
\label{tab:astabench_tasks}
\end{table*}

Each AstaBench task requires an agent to complete an end-to-end
research cycle involving experimental implementation, execution,
result analysis, and technical report generation. We use the official
task-specific rubrics and LLM-as-judge evaluation pipeline, with
MiniMax-M3 serving as the evaluator.

For every rubric criterion, the evaluator independently examines
three output facets: the generated report, the generated code, and
the supporting artifacts, including datasets, execution logs, figures,
and model outputs. Each facet is classified as
\textit{meets criterion}, \textit{fails criterion}, or
\textit{no evidence either way}. The facet-level assessments are then
combined into a binary verdict for the corresponding criterion. The
official task score is the arithmetic mean of the binary criterion
verdicts, multiplied by 100 for presentation.

In addition to the mean evaluation score, we report the number of
successfully completed tasks. A task is considered successfully
completed when the agent run terminates without an execution-level
failure and produces a valid submission that can be processed and
scored by the official evaluator.

\paragraph{Models and agent harnesses.}
We evaluate generalization along two complementary axes on
ResearchClawBench. In cross-model evaluation, we hold the Codex agent
harness fixed and vary the backbone LLM among GPT-5.4, GLM-5.2, and MiniMax-M3. In cross-harness evaluation, we hold the DeepSeek-V4-Flash backbone fixed and vary the surrounding agent implementation among Claude Code, OpenClaw, and OpenScience.

For AstaBench, we evaluate three representative configurations:
Claude Code with DeepSeek-V4-Flash, OpenClaw with
DeepSeek-V4-Flash, and Codex with GPT-5.4-mini. The complete
configurations and corresponding benchmark evaluators are summarized
in Table~\ref{tab:model_harness_configs}.

\begin{table*}[t]
\centering
\small
\setlength{\tabcolsep}{5pt}
\renewcommand{\arraystretch}{1.08}
\begin{tabular}{
    p{0.21\textwidth}
    p{0.18\textwidth}
    p{0.21\textwidth}
    p{0.24\textwidth}
    p{0.12\textwidth}
}
\toprule
\textbf{Benchmark}
& \textbf{Setting}
& \textbf{Agent harness}
& \textbf{Backbone LLM}
& \textbf{Evaluator}
\\
\midrule

ResearchClawBench
& Cross-model
& Codex
& GPT-5.4
& GPT-5.1
\\

ResearchClawBench
& Cross-model
& Codex
& GLM-5.2
& GPT-5.1
\\

ResearchClawBench
& Cross-model
& Codex
& MiniMax-M3
& GPT-5.1
\\

ResearchClawBench
& Cross-harness
& Claude Code
& DeepSeek-V4-Flash
& GPT-5.1
\\

ResearchClawBench
& Cross-harness
& OpenClaw
& DeepSeek-V4-Flash
& GPT-5.1
\\

ResearchClawBench
& Cross-harness
& OpenScience
& DeepSeek-V4-Flash
& GPT-5.1
\\

\midrule

AstaBench
& Transfer
& Claude Code
& DeepSeek-V4-Flash
& MiniMax-M3
\\

AstaBench
& Transfer
& OpenClaw
& DeepSeek-V4-Flash
& MiniMax-M3
\\

AstaBench
& Transfer
& Codex
& GPT-5.4-mini
& MiniMax-M3
\\

\bottomrule
\end{tabular}
\caption{
Model, agent-harness, and evaluator configurations used in the
ResearchClawBench and AstaBench experiments.
}
\label{tab:model_harness_configs}
\end{table*}

For every paired comparison, the \textit{vanilla} setting denotes the
original agent without {\our}. The \(+{\our}\) setting uses the same
backbone model, agent harness, task instruction, task-visible data,
workspace, and available tools, while additionally performing rubric
induction, criterion-level verification, and targeted revision.
Accordingly, the paired configurations are controlled for the
underlying agent and execution environment, but are not compute-matched,
because {\our} introduces the additional model calls required by the
method.

\paragraph{Execution and evaluation protocol.}
Each task--configuration pair is executed once, producing one
scientific report and one set of supporting artifacts. Each resulting
submission is then independently evaluated three times using the same
benchmark evaluator and evaluation protocol. The arithmetic mean of
the three judge scores is used as the task-level score.

For ResearchClawBench, the overall score is computed as the arithmetic
mean over all 40 task-level scores, and each domain score is computed
over its four constituent tasks. For AstaBench, the mean score is
computed over the fixed 20-task subset. Successful-task counts are
computed independently from the rubric-based evaluation scores.

Within each paired comparison, the \textit{vanilla} and \(+{\our}\)
settings receive the same task inputs, task-visible resources,
workspace, and tool access. Available tools include the native
capabilities of the corresponding agent harness, such as file
inspection, code execution, and web search, as well as the external
retrieval services used by {\our}. The hidden target paper and the
official benchmark evaluation rubric remain inaccessible to both
settings during research execution.

We do not impose a shared explicit cap on tokens, tool calls,
wall-clock runtime, or monetary cost across heterogeneous agent
harnesses. Each system instead operates under the execution, context,
and tool constraints of its native environment. Our primary evaluation
therefore focuses on output quality rather than budget-normalized
efficiency.

Rubric-guided verification and revision use a maximum of three
revision rounds with adaptive early stopping. To distinguish the
effect of structured rubric guidance from that of additional
opportunities for revision, we additionally evaluate a three-round
rubric-free self-refinement baseline starting from the same initial
reports and using the same underlying research agent. The complete
verification, revision, early-stopping, and self-refinement protocols
are described in the following subsection.

\subsection{Rubric Quality Evaluation}
\label{app:rubric_evaluation}

\paragraph{Compared rubric variants and scoring protocol.}
For each of the 40 ResearchClawBench tasks, we evaluate two rubric
variants: the \textit{instruction-derived rubric skeleton} produced
after Rubric Skeleton Induction, and the \textit{task-specific
executable rubric} produced after Scientific Literature Grounding,
Task-Data Exploration, and Criterion Synthesis.

Each rubric is evaluated along four dimensions:
Specificity, Actionability, Evidence Verifiability, and Scientific
Core Coverage. All dimensions use a five-point ordinal scale, where
a higher value indicates stronger rubric quality. We use MiniMax-M3 as the rubric-quality judge for all evaluations. To reduce cross-dimensional interference, we use four standalone judge prompts, each of which evaluates exactly one dimension. The judge receives only one candidate rubric in each call and is explicitly instructed not to score the other three dimensions.

For every judge call, the evaluator is provided with the task
instruction, a description of the task-visible data, the candidate
rubric, and the official benchmark checklist. The checklist is used
only as evaluator-side evidence for identifying the intended
scientific success conditions. It is never exposed to the agent during
rubric construction, research execution, verification, or revision.
The judge is instructed to assess semantic and scientific equivalence
rather than lexical overlap with the checklist.

The rubric skeleton and executable rubric are evaluated in separate calls. The judge is not provided with the generated report or the alternative rubric variant. The overall quality score of rubric $\mathcal{R}$ is computed as the unweighted arithmetic mean of the four dimension scores:
\[
Q(\mathcal{R})
=
\frac{1}{4}
\left(
s_{\mathrm{spec}}
+
s_{\mathrm{act}}
+
s_{\mathrm{ver}}
+
s_{\mathrm{core}}
\right),
\]
where each score is an integer between 1 and 5.

\paragraph{Evaluation dimensions.}

\textbf{Specificity.}
Specificity measures how precisely the rubric states what is to be
done and produced. A specific rubric names the relevant methods or
experiments, variables, conditions, metrics, comparators, thresholds,
and expected artifacts at a level that makes the intended analysis
unambiguous. Specificity concerns the concreteness of the
specification; it does not by itself establish that the plan is
executable, scientifically valid, or directed at the correct
scientific objective.

\textbf{Actionability.}
Actionability measures whether a competent agent can execute the
rubric using the available inputs without inventing consequential
parts of the plan. An actionable rubric provides sufficient
operational steps, dependencies, parameter or method choices,
decision rules, and expected outputs to support implementation.
Actionability does not reward detail that is irrelevant to execution
and does not guarantee that an executable plan addresses the task's
scientific core.

\textbf{Evidence Verifiability.}
Evidence Verifiability measures whether the rubric requires
inspectable and traceable evidence that permits independent
verification of both task completion and the resulting scientific
claims. Verifiable evidence may include quantitative tables, figures,
machine-readable outputs, controls, baseline comparisons, uncertainty
estimates, provenance, and explicit links between claims and generated
results. This dimension evaluates the checkability and diagnostic
value of the evidence rather than the number of artifacts requested.

\textbf{Scientific Core Coverage.}
Scientific Core Coverage measures the extent to which the rubric
faithfully covers and operationalizes the task's core scientific
success conditions. It evaluates whether the rubric's objectives,
methods or scientifically equivalent alternatives, metrics,
comparisons, artifacts, scope, conventions, evidence, and required
conclusions collectively target the intended scientific problem.
Scores decrease when important success conditions are omitted or
underweighted, when the rubric drifts toward an adjacent problem, or
when core goals are replaced by non-equivalent proxies. Alternative
methods and artifacts receive full credit when they provide
scientifically equivalent or stronger evidence for the same core
claim.

The four dimensions capture complementary properties. A rubric may
be highly specific and actionable but receive a low Scientific Core
Coverage score if it precisely operationalizes the wrong analysis.
Conversely, a concise rubric may identify the correct scientific core
but receive lower Specificity or Actionability scores if important
implementation choices remain unspecified. Evidence Verifiability is
assessed separately: a scientifically aligned and executable plan may
still be difficult to verify when it does not require inspectable
outputs that directly support its conclusions.

\paragraph{Complete rubric-quality scoring prompts.}
The following four prompts are used independently. In the JSON
templates, the value of \texttt{score} is replaced with an integer
from 1 to 5. Rubric item identifiers refer to goal IDs for rubric skeletons and criterion IDs for executable rubrics.

\clearpage
\onecolumn

\noindent
{\large\bfseries Complete Rubric-Quality Scoring Prompts}

The following four prompts independently evaluate Specificity, Actionability, Evidence Verifiability, and Scientific Core Coverage.
Each judge call receives one candidate rubric and evaluates exactly one dimension.

\begin{PromptBox}{Specificity Scoring Prompt}
You are a senior scientific-methods reviewer. Evaluate exactly one construct: SPECIFICITY. 

Definition:
Specificity is the precision with which the rubric states what is to be done and produced. It concerns whether the relevant methods or experiments, variables, conditions, metrics, comparators, thresholds, and expected artifacts are concrete enough to make the intended analysis unambiguous.

Evaluation rules:
1. Inspect the rubric's core items rather than rewarding length or technical jargon.
2. Credit concrete methods, conditions, variables, metrics, comparisons, thresholds, named outputs, and artifact requirements when they remove genuine ambiguity.
3. Do not reward irrelevant detail or copied wording from the task or benchmark checklist.
4. Do not lower the score merely because the specified plan may be difficult to execute, scientifically questionable, weakly evidenced, or misdirected; those properties belong to other dimensions.
5. Cite rubric item IDs and identify the most consequential specified and unspecified details.

Score anchors:
1 = The rubric is almost entirely generic; core methods, conditions, metrics, and outputs are not concretely specified.
2 = A few concrete details are present, but most consequential aspects remain ambiguous.
3 = The main analysis is identifiable, but important conditions, metrics, comparisons, thresholds, or artifacts are underspecified.
4 = Nearly all core details are explicit; only minor ambiguities remain.
5 = The core analysis is precise and unambiguous across methods, conditions, variables, metrics, comparisons, thresholds, and expected artifacts.

Return JSON only. Set "score" to an integer from 1 to 5:
{
  "dimension": "specificity",
  "score": 1,
  "evidence": "Concise justification citing rubric item IDs.",
  "specified_strengths": ["..."],
  "remaining_ambiguities": ["..."]
}

TASK INSTRUCTION:
[task_instruction]

AVAILABLE-DATA DESCRIPTION:
[available_data]

BENCHMARK CHECKLIST -- EVALUATOR-ONLY CONTEXT:
[benchmark_checklist]

RUBRIC TO SCORE:
[rubric]
\end{PromptBox}

\begin{PromptBox}{Actionability Scoring Prompt}
You are a senior scientific-methods reviewer. Evaluate exactly one construct: ACTIONABILITY. 

Definition:
Actionability is the degree to which a competent agent can execute the rubric using the available inputs without inventing consequential parts of the plan. It concerns operational steps, dependencies, method and parameter choices, decision rules, and deliverables that reduce implementation uncertainty.

Evaluation rules:
1. Determine whether the rubric supplies an executable route from the available data to the required outputs.
2. Identify consequential methods, parameters, dependencies, preprocessing choices, decision rules, or deliverables that the agent would otherwise have to invent.
3. Credit flexibility when multiple choices are genuinely interchangeable and the rubric supplies a sound decision rule or acceptable range.
4. Do not reward detail that does not help execution. Do not judge whether the chosen scientific target is the correct one; that belongs to Scientific Core Coverage.
5. Cite rubric item IDs and focus on implementation consequences.

Score anchors:
1 = No executable route is supplied; the agent must design nearly the entire plan.
2 = Major methods, dependencies, parameters, or decisions must be invented.
3 = An executable outline exists, but some consequential implementation choices remain unresolved.
4 = The core workflow is directly executable, with only minor or low-impact choices left open.
5 = An end-to-end operational plan specifies all consequential steps, dependencies, decision rules, and outputs needed for execution.

Return JSON only. Set "score" to an integer from 1 to 5:
{
  "dimension": "actionability",
  "score": 1,
  "evidence": "Concise justification citing rubric item IDs.",
  "executable_elements": ["..."],
  "missing_decisions": ["..."]
}

TASK INSTRUCTION:
[task_instruction]

AVAILABLE-DATA DESCRIPTION:
[available_data]

BENCHMARK CHECKLIST -- EVALUATOR-ONLY CONTEXT:
[benchmark_checklist]

RUBRIC TO SCORE:
[rubric]
\end{PromptBox}

\begin{PromptBox}{Evidence Verifiability Scoring Prompt}
You are a senior scientific-methods reviewer. Evaluate exactly one construct: EVIDENCE VERIFIABILITY.

Definition:
Evidence Verifiability is the degree to which the rubric requires inspectable and traceable evidence that allows an independent reviewer to verify task completion and evaluate whether the scientific claims follow from the generated results.

Evaluation rules:
1. Inspect required quantitative tables, figures, machine-readable outputs, controls, baseline comparisons, uncertainty estimates, provenance, and claim-to-evidence links when scientifically applicable.
2. Judge the diagnostic and confirmatory value of evidence, not the number of files.
3. Credit explicit must-show criteria, metrics, uncertainty, comparison targets, and traceable paths that make an artifact independently checkable.
4. Penalize rubrics that allow conclusions to be asserted without supporting results, or request decorative outputs that cannot test the claims.
5. Do not judge whether the rubric targets the correct scientific goal; that belongs to Scientific Core Coverage.

Score anchors:
1 = Conclusions can be asserted without any inspectable supporting evidence.
2 = Outputs are requested only vaguely or lack checkable criteria and claim links.
3 = Some core results are inspectable, but important claims, comparisons, uncertainty, or diagnostics remain weakly supported.
4 = Core claims are supported by checkable quantitative artifacts and appropriate diagnostics; only minor traceability gaps remain.
5 = The evidence chain is complete and traceable from method to result to claim, including controls, uncertainty, comparisons, and provenance where relevant.

Return JSON only. Set "score" to an integer from 1 to 5:
{
  "dimension": "evidence_verifiability",
  "score": 1,
  "evidence": "Concise justification citing rubric item IDs.",
  "verifiable_artifacts": ["..."],
  "verification_gaps": ["..."]
}

TASK INSTRUCTION:
[task_instruction]

AVAILABLE-DATA DESCRIPTION:
[available_data]

BENCHMARK CHECKLIST -- EVALUATOR-ONLY CONTEXT:
[benchmark_checklist]

RUBRIC TO SCORE:
[rubric]
\end{PromptBox}

\begin{PromptBox}{Scientific Core Coverage Scoring Prompt}
You are a senior scientific-methods reviewer. Evaluate exactly one construct: SCIENTIFIC CORE COVERAGE.

Definition:
Scientific Core Coverage is the extent to which the rubric faithfully covers and operationalizes the task's core scientific success conditions through aligned objectives, scientifically equivalent methods, metrics, comparisons, artifacts, scope, conventions, evidence, and required conclusions. It excludes material omission, goal drift, and substitution by adjacent but non-equivalent proxies.

Evaluation rules:
1. Infer the task's core scientific success conditions from the instruction and available data. Use the benchmark checklist and its criterion weights, when available, only as evaluator-side evidence of importance.
2. Map each core condition to rubric items and classify it as aligned, scientifically equivalent, partial, proxy, missing, or conflicting.
3. Judge semantic and scientific-function equivalence, never wording overlap. Alternative methods and artifacts deserve full credit when they provide equivalent or stronger evidence for the same core claim.
4. Penalize omission or underweighting of high-importance goals, goal drift, non-equivalent proxy substitution, artifact or metric mismatch, scope dilution, incompatible conventions, and  conclusion anchoring.
5. Missing low-level implementation detail alone belongs to Specificity or Actionability and must not reduce this score when the scientific target remains intact.

Score anchors:
1 = The rubric operationalizes a substantially different problem, contradicts a core success condition, or is dominated by non-equivalent proxies.
2 = Major omissions, conflicts, or proxy substitutions affect one or more core or high-importance success conditions.
3 = The main objective is recognizable, but an important success condition is missing, weakened, underweighted, or replaced.
4 = Core success conditions are faithfully operationalized; only minor or peripheral mismatches remain, and valid equivalent alternatives are used appropriately.
5 = All core success conditions are tightly covered by aligned or scientifically equivalent objectives, methods, metrics, comparisons, artifacts, scope, and evidence, with no material drift, omission, or proxy replacement.

Return JSON only. Set "score" to an integer from 1 to 5. The allowed values of "importance" are "core", "important", and "peripheral". The allowed values of "status" are "aligned", "equivalent", "partial", "proxy", "missing", and "conflicting":
{
  "dimension": "scientific_core_coverage",
  "score": 1,
  "evidence": "Concise justification citing rubric item IDs and core consequences.",
  "criterion_audit": [
    {
      "criterion": "...",
      "importance": "core",
      "status": "aligned",
      "rubric_items": ["R1"],
      "rationale": "..."
    }
  ],
  "core_gaps_or_drift": ["..."]
}

TASK INSTRUCTION:
[task_instruction]

AVAILABLE-DATA DESCRIPTION:
[available_data]

BENCHMARK CHECKLIST -- EVALUATOR-ONLY CONTEXT:
[benchmark_checklist]

RUBRIC TO SCORE:
[rubric]
\end{PromptBox}

\clearpage
\twocolumn

\subsection{Rubric-Guided Revision Evaluation}
\label{app:revision_evaluation}

\paragraph{Experimental configuration and shared initialization.}
We compare rubric-guided revision with rubric-free holistic
self-refinement on all 40 ResearchClawBench tasks. All checkpoint-0
reports, verification calls, targeted revisions, and rubric-free
revisions are executed with OpenClaw using
\texttt{deepseek/deepseek-v4-flash}. There is no separately instantiated
verifier model. Instead, the same OpenClaw agent invokes the
\texttt{report-verifier} skill within the task session.

Both revision strategies start from the same set of 40 frozen
checkpoint-0 reports and figures. These checkpoint-0 outputs are
generated under the task-specific executable rubric after planning,
literature grounding, experiment design, experiment execution, figure
generation, and initial report writing, but before any verifier call
or verifier-driven revision. Their mean ResearchClawBench score is
18.309083, reported as 18.31 in the main paper.

For the rubric-free condition, the canonical checkpoint-0
\texttt{report.md} file and its figures are copied into a new isolated
workspace. SHA-256 hashes are computed before and after copying to
verify that the starting report and figures are unchanged. We do not
claim that the two strategies share an identical complete workspace:
checkpoint-0 code and output directories are not frozen or copied
because they may contain rubric-related state and later feedback.
Instead, the rubric-free workspace is supplied again with the original
instruction, task data, and related work, and the agent may create new
code, outputs, and analyses during refinement.

\paragraph{Criterion-level verification.}
Before each guided revision, the current artifact is evaluated by the
same OpenClaw agent through a verification-only call. The verifier
examines the current report, generated experiments and results,
figures and tables, quantitative values, conclusions, and
mechanistic or analytical explanations. It also checks whether the
artifacts required by the task-specific executable rubric are present
or explicitly represented in the report, including their filenames,
titles, captions, display names, and detailed descriptions.

The verifier is instructed to distinguish generated experimental
evidence from statements supported only by prior literature.
Conclusions are considered adequately supported only when they follow
from evidence produced within the current task execution.

For each scientific goal, the verifier records whether the report
contains the required experiment, results, figure or table,
conclusion, and mechanism or analytical explanation. It also evaluates
individual rubric items and produces stable feedback items specifying
the remaining issue and the required corrective action.

The top-level \texttt{overall\_pass} field is an LLM judgment rather
than a deterministic conjunction of all criterion-level Boolean
values. An artifact passes when all scientific goals have sufficient
experimental support, interpretation, and conclusion coverage, and
all high-priority rubric items are either satisfied or explicitly and
scientifically justified by the generated evidence. Lower-priority
items are not required to pass individually. The verifier does not
produce a separate structured \texttt{not\_applicable} state; an
inapplicable requirement may instead be justified in free text while
the corresponding structured field remains Boolean.

The expected verifier output follows the schema below.

\begin{PromptBox}{Verifier Output Schema}
{
  "source": "rcbench_report_verifier",
  "overall_pass": false,
  "goal_checks": [
    {
      "goal_id": "G1",
      "pass": false,
      "coverage": {
        "experiment": true,
        "results": true,
        "figure_or_table": false,
        "conclusion": true,
        "mechanism_or_analysis": false
      },
      "missing": ["..."],
      "fixes": ["..."]
    }
  ],
  "rubric_item_checks": [
    {
      "rubric_item_id": "R1",
      "pass": false,
      "missing": ["..."],
      "fixes": ["..."]
    }
  ],
  "feedback_items": [
    {
      "feedback_id": "F001",
      "goal_id": "G1",
      "rubric_item_id": "R1",
      "issue": "...",
      "required_action": "..."
    }
  ],
  "artifact_checks": [
    "path/name, present?, described?, issue"
  ],
  "required_iterations": ["..."]
}
\end{PromptBox}

The runner reads the top-level \texttt{overall\_pass} value directly
and does not independently reconstruct the pass decision from the
individual checks. If the verifier output is missing or cannot be
parsed, the runner treats the artifact as not passing and proceeds to
revision. No separate JSON-schema validation or verifier-specific
retry is performed. A task run is marked as failed only when the
OpenClaw process exits with a nonzero status.

\paragraph{Targeted revision and early stopping.}
If verification does not pass, the current
\texttt{report\_rubric\_check.json} is supplied to the same
OpenClaw agent for one targeted revision. The revision call has access
to the current workspace, \texttt{INSTRUCTIONS.md}, the current
report, the executable rubric, goal-level experiment plans, existing
code, outputs, figures, and the current verifier feedback.

The agent may modify code, rerun or add experiments and analyses,
regenerate figures or tables, correct numerical results, revise
unsupported claims, and update the complete
\texttt{report/report.md}. The revision call is instructed not to
invoke the verifier itself; verification and revision are executed as
separate agent calls.

We allow at most three revision rounds. The resulting execution
sequence is
\[
V^{(0)}
\rightarrow R^{(1)}
\rightarrow V^{(1)}
\rightarrow R^{(2)}
\rightarrow V^{(2)}
\rightarrow R^{(3)}
\rightarrow V^{(3)},
\]
where \(V^{(t)}\) denotes verification of checkpoint \(t\), and
\(R^{(t)}\) denotes the revision producing checkpoint \(t\).
Consequently, each task receives at most three revision calls and four
verification calls. The procedure terminates immediately when a
verification call returns \texttt{overall\_pass=true}. If the artifact
does not pass earlier, a final verification is performed after the
third revision.

\paragraph{Adaptive stopping and checkpoint aggregation.}
Rubric-guided revision generates and evaluates only reports that are
actually produced. Across the 40 tasks, the experiment contains 103
distinct report checkpoints: 40 checkpoint-0 reports, 40 checkpoint-1
reports, 17 checkpoint-2 reports, and 6 checkpoint-3 reports.

When a task passes verification, no artificial reports are generated
for later revision budgets. Instead, its most recent real report and
its previously obtained score are carried forward when computing
later checkpoint-level means. The carried-forward report is not
rerun or rescored. Thus, every reported guided checkpoint mean is
computed over all 40 tasks, while only tasks that have not yet stopped
produce a new report at the next revision round.

Each actually generated report checkpoint is independently scored
three times by the GPT-5.1 ResearchClawBench judge, and the arithmetic
mean of the three scores is used as its task-level score.

\paragraph{Rubric-free holistic self-refinement.}
The rubric-free baseline starts from the same frozen checkpoint-0
report and figures and uses the same OpenClaw and
\texttt{deepseek/deepseek-v4-flash} configuration. It is explicitly
prohibited from accessing, constructing, inferring, or searching for
the task-specific executable rubric, the hidden benchmark checklist,
the target study, rubric scores, or prior verifier feedback.

At each round, the agent invokes the
\texttt{holistic-self-review} skill and performs a holistic
scientific review followed immediately by one revision within the
same agent call. The agent may use tools, write code, rerun or add
experiments and analyses, modify figures and tables, revise results
and claims, and update the complete report.

Unlike rubric-guided revision, rubric-free self-refinement does not
use verifier-based early stopping. All 40 tasks execute three
review--revision cycles, producing 40 real reports at each of
checkpoints 1, 2, and 3. In the first round, the runner requires
between one and six prioritized issues. In the second and third
rounds, an empty issue list and a no-op revision are allowed when no
substantive deficiency is identified. Each resulting report is
independently scored three times by the same GPT-5.1 judge.

This baseline controls for repeated opportunities to inspect and
revise the report, but it is not a strictly compute-matched or
workspace-identical control. Guided revision separates verification
and revision into two calls and uses adaptive stopping, whereas
rubric-free refinement combines review and revision into one call and
always performs three rounds. Guided revision also retains the
original code and output state, while the rubric-free condition starts
from the frozen report and figures in a newly initialized workspace.

\clearpage
\onecolumn

\section{Rubric Output Formats}
\label{app:rubric_formats}

This section presents the output formats of the
instruction-derived rubric skeleton and the task-specific executable
rubric. One complete entry is shown for each format; the remaining
entries are omitted for brevity.

\paragraph{Instruction-derived rubric skeleton.}
The rubric skeleton represents the task as a set of atomic scientific
goals. Each goal contains a unique identifier, a concise title, and a
concrete task description.

\begin{PromptBox}{Instruction-Derived Rubric Skeleton}
{
  "goals": [
    {
      "goal_id": "G1",
      "title": "Characterize spatial concentration of cloud-seeding projects",
      "task": "Analyze the geographic distribution of cloud-seeding projects across U.S. states using the published records, and identify which states account for the majority of reported activity."
    },
    {
      "...": "Additional scientific goals are omitted."
    }
  ]
}
\end{PromptBox}

\paragraph{Task-specific executable rubric.}
The executable rubric augments the scientific goals with concrete
experiments, required evidence artifacts, metrics, comparisons, data
sources, expected conclusions, priorities, and execution constraints.

\begin{PromptBox}{Task-Specific Executable Rubric}
{
  "source": "rcbench_experiments_design",
  "task_summary": "Reproduce and validate the central empirical conclusions from a published NOAA weather-modification dataset of U.S. cloud seeding projects (2000-2025). Generate spatial concentration maps, annual activity dynamics, purpose composition, and agent-apparatus deployment patterns using transparent, script-based analysis.",
  "selected_environment": ".venv-rcb-base",
  "environment_reason": "The selected environment provides the packages required for tabular analysis, statistical computation, and geospatial visualization.",
  "selected_strategy": "Independently compute the required statistics and figures from the structured dataset and synthesize the results into an empirical assessment.",

  "rubric_items": [
    {
      "id": "R1",
      "linked_goal_ids": [
        "G1"
      ],
      "goal": "Quantify state-level spatial concentration of cloud seeding projects and produce a choropleth map showing geographic distribution.",
      "experiment_plan_path": "outputs/goals/G1/experiment_plan.md",
      "experiments_to_run": [
        "Group records by state and count projects",
        "Compute rank, percentage share, and Gini coefficient",
        "Merge state counts with GeoJSON and generate a choropleth map",
        "Create a state-by-purpose heatmap"
      ],
      "required_artifacts": [
        {
          "type": "figure",
          "name": "spatial_distribution.png",
          "expected_path": "report/images/spatial_distribution.png",
          "must_show": "A choropleth map of U.S. states colored by the number of cloud seeding projects, showing strong western concentration and identifying California, Colorado, Texas, and Utah among the leading states.",
          "description_requirement": "Explain the geographic pattern of cloud seeding activity and the factors associated with western concentration."
        },
        {
          "type": "figure",
          "name": "state_purpose_heatmap.png",
          "expected_path": "report/images/state_purpose_heatmap.png",
          "must_show": "A state-by-purpose heatmap demonstrating how project purposes vary across states.",
          "description_requirement": "Describe how project purposes vary across states and regions."
        },
        {
          "type": "table",
          "name": "top_states_table",
          "expected_path": "report/report.md",
          "must_show": "A ranked list of the leading states with project counts and percentage shares.",
          "description_requirement": "Summarize the state-level concentration statistics."
        }
      ],
      "metrics_or_values": [
        "Projects per state",
        "Percentage share of the top three and top five states",
        "Gini coefficient"
      ],
      "baselines_or_comparisons": [
        "Expected geographical concentration in water-stressed western states"
      ],
      "data_sources": [
        "data/dataset1_cloud_seeding_records/cloud_seeding_us_2000_2025.csv[state]",
        "data/dataset1_cloud_seeding_records/us_states.geojson"
      ],
      "expected_conclusions": [
        "Cloud seeding is heavily concentrated in western U.S. states",
        "California has the highest project count",
        "The spatial distribution reflects water-resource management priorities"
      ],
      "priority": "high",
      "risk_if_missing": "Loss of the primary spatial evidence supporting geographic concentration claims."
    },
    {
      "...": "Additional executable rubric items are omitted."
    }
  ],

  "execution_plan": [
    {
      "...": "Task-level execution steps are omitted."
    }
  ],

  "claims_to_avoid": [
    "Do not claim causal relationships between cloud seeding and precipitation changes.",
    "... additional unsupported claims are omitted ..."
  ]
}
\end{PromptBox}

\end{document}